\documentclass[10pt,twocolumn,letterpaper]{article}

\usepackage[pagenumbers]{cvpr} 

\usepackage{graphicx}
\usepackage{amsmath}
\usepackage{amssymb}
\usepackage{booktabs}
\usepackage{multirow}
\usepackage{makecell}
\usepackage[dvipsnames]{xcolor}

\usepackage[pagebackref,breaklinks,colorlinks]{hyperref}

\usepackage[capitalize]{cleveref}
\crefname{section}{Sec.}{Secs.}
\Crefname{section}{Section}{Sections}
\Crefname{table}{Table}{Tables}
\crefname{table}{Tab.}{Tabs.}

\newcommand{\xhdr}[1]{\vspace{0pt} \noindent {\textbf{#1}}}

\newcommand{\oursloss}{UniCL}
\newcommand{\ourslossfull}{Unified Contrastive Learning}

\usepackage{amsmath}
\usepackage{amsfonts}
\usepackage{amssymb}
\usepackage{wrapfig}
\usepackage{subcaption}
\usepackage{multirow}
 \usepackage{mathtools} 

\usepackage{verbatim}

\usepackage{anyfontsize}

\usepackage{microtype}
\usepackage{graphicx}
\usepackage{booktabs} 
\usepackage{multirow}
\usepackage{amsmath,amssymb}
\usepackage{booktabs}
\usepackage{caption,subcaption}
\usepackage{xcolor}
\usepackage{enumitem}
\setitemize{itemsep=10pt,topsep=0pt,parsep=0pt,partopsep=0pt}
\usepackage{adjustbox}

\newcommand{\RN}[1]{%
	\textup{\lowercase\expandafter{\it \romannumeral#1}}%
}
\usepackage{tabu}

\newcommand{\beq}{\vspace{0mm}\begin{equation}}
\newcommand{\eeq}{\vspace{0mm}\end{equation}}
\newcommand{\beqs}{\vspace{0mm}\begin{eqnarray}}
\newcommand{\eeqs}{\vspace{0mm}\end{eqnarray}}
\newcommand{\barr}{\begin{array}}
\newcommand{\earr}{\end{array}}

\newcommand{\Umat}[0]{{{\bf U}}}
\newcommand{\Vmat}[0]{{{\bf V}}}
\newcommand{\Wmat}[0]{{{\bf W}}}

\newcommand{\tv}[0]{{\boldsymbol{t}}}
\newcommand{\uv}{\boldsymbol{u}}
\newcommand{\vv}{\boldsymbol{v}}
\newcommand{\wv}{\boldsymbol{w}}
\newcommand{\xv}{\boldsymbol{x}}

\newcommand{\thetav}{\boldsymbol{\theta}}

\newcommand{\phiv}{\boldsymbol{\phi}}

\newcommand{\R}{\mathbb{R}}

\newcommand{\Xcal}{\mathcal{X}}
\newcommand{\Ycal}{\mathcal{Y}}

\newcommand{\Bcal}{\mathcal{B}}

\newcommand{\Tcal}{\mathcal{T}}

\newcommand{\Lcal}{\mathcal{L}}

\newcommand{\Pcal}{\mathcal{P}}

\newcommand{\Scal}{\mathcal{S}}

\usepackage{color, colortbl}
\definecolor{Gray}{gray}{0.93}

\usepackage{xcolor,amsmath}
\usepackage[linesnumbered,ruled,vlined]{algorithm2e}
\DontPrintSemicolon

\SetKwComment{Comment}{\color{green!50!black}\# }{}

\newcommand{\var}{\texttt}
\newcommand{\FuncCall}[2]{\texttt{\bfseries #1(#2)}}
\SetKwProg{Function}{def}{:}{}

\SetKwProg{For}{for}{:}{}
\SetKwProg{If}{if}{:}{}

\renewcommand{\comment}[1]{} 

\def\cvprPaperID{2347} 
\def\confName{CVPR}
\def\confYear{2022}

\begin{document}

\title{Unified Contrastive Learning in Image-Text-Label Space}
\author{
  Jianwei Yang$^{1}$\thanks{equal contribution} \,\, Chunyuan Li$^{1*}$ \,\, Pengchuan Zhang$^{1*}$ \,\, Bin Xiao$^{2*}$ \\ 
  {Ce Liu}$^{2}$ \,\, {Lu Yuan}$^{2}$ \,\, {Jianfeng Gao}$^{1}$ \\
  $^1$Microsoft Research at Redmond, $^2$Microsoft Cloud + AI\\
  \texttt{\{jianwyan,chunyl,penzhan,bixi,liuce,luyuan,jfgao\}@microsoft.com} \\
}
\maketitle

\begin{abstract} 
Visual recognition is recently learned via either \emph{supervised learning} on human-annotated image-label data or \emph{language-image} contrastive learning with webly-crawled image-text pairs. While supervised learning may result in a more discriminative representation, language-image pretraining shows unprecedented zero-shot recognition capability, largely due to the different properties of data sources and learning objectives. In this work, we introduce a new formulation by combining the two data sources into a common image-text-label space. In this space, we propose a new learning paradigm, called \emph{Unified Contrastive Learning (\oursloss)} with a single learning objective to seamlessly prompt the synergy of two data types. Extensive experiments show that our \oursloss{} is an effective way of learning semantically rich yet discriminative representations, universally for image recognition in zero-shot, linear-probing, fully finetuning and transfer learning scenarios. Particularly, it attains gains up to \textbf{9.2}\% and \textbf{14.5}\% in average on zero-shot recognition benchmarks over the language-image contrastive learning and supervised learning methods, respectively. In linear probe setting, it also boosts the performance over the two methods by \textbf{7.3}\% and \textbf{3.4}\%, respectively. Our study also indicates that \oursloss{} stand-alone is a good learner on pure image-label data, rivaling the supervised learning methods across three image classification datasets and two types of vision backbones, ResNet and Swin Transformer. Code is available at: \url{https://github.com/microsoft/UniCL}.
\end{abstract}

\section{Introduction}

\begin{figure}[t]
	\centering
	\includegraphics[width=0.80\linewidth]{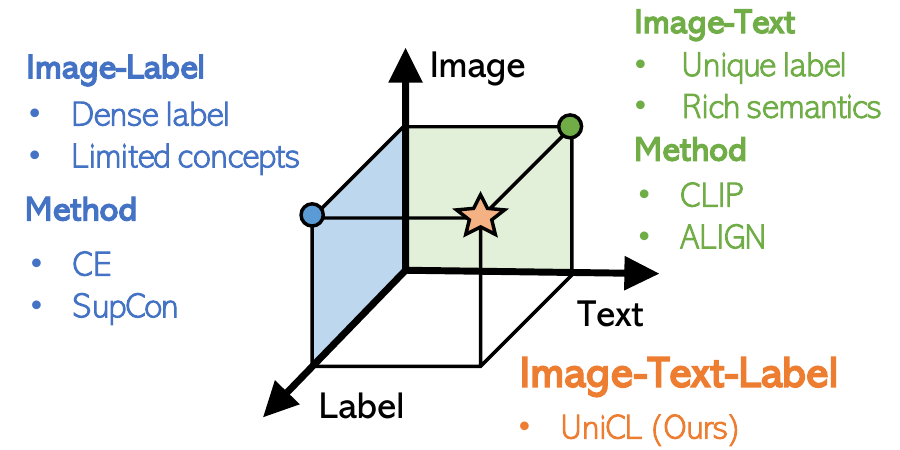}
	\includegraphics[width=0.99\linewidth]{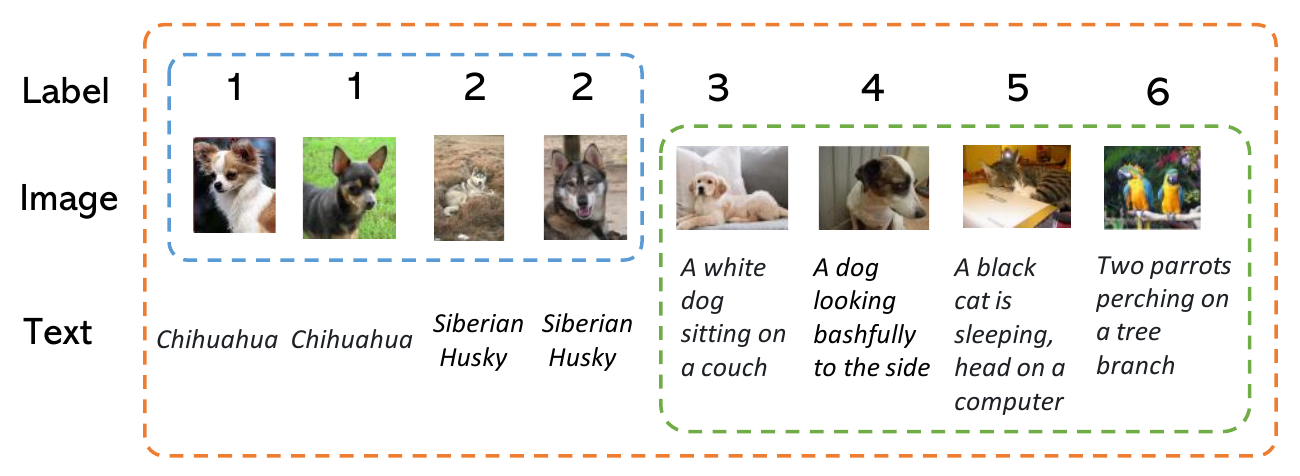}
    \vspace{-3pt}
    \caption{Unified contrastive learning paradigm in the \textcolor{orange}{\bf image-text-label} space, which recovers the supervised learning (\textit{e.g.}, Cross-Entropy (CE)~\cite{murphy2012machine} or Supervised Contrastive Learning (SupCon)~\cite{khosla2020supervised}) on \textcolor{RoyalBlue}{\bf image-label} data, and language-image contrastive learning (\textit{e.g.}, CLIP~\cite{radford2021learning} or ALIGN~\cite{jia2021scaling}) on \textcolor{LimeGreen}{\bf image-text} data.}
    \label{fig:teaser}
    \vspace{-3mm}
\end{figure}

Learning to recognize visual concepts in an image has been a fundamental and long-standing research problem. Typically, this can be tackled via either supervised learning on human-annotated image-label pairs~\cite{deng2009imagenet} or contrastive learning on webly-crawed image-text pairs~\cite{radford2021learning,jia2021scaling}. When fueled with clean and large-scale human-annotated image-label data, \textit{e.g.}, ImageNet~\cite{deng2009imagenet}, supervised learning can attain decent visual recognition capacities over the given categories~\cite{krizhevsky2012imagenet,szegedy2016rethinking,he2016deep} and also powerful transfer learning abilities~\cite{donahue2014decaf,kolesnikov2020big}. Nevertheless, collecting precise image-label data can be a laborious and expensive process, not to say its difficulty to scale up to numerous visual concepts\footnote{The largest scale but private JFT-300M covers 18,291 concepts.}. On the other hand, language-image contrastive learning has recently emerged as a promising approach by leveraging huge amounts of webly-crawled image-text pairs. These pairs are usually noisy, free-form but cover lots of visual concepts. As demonstrated in CLIP~\cite{radford2021learning} and ALIGN~\cite{jia2021scaling}, models learned from hundreds of millions of image-text pairs can attain impressive low-shot recognition performance for a wide range of visual understanding scenarios. 
Though these image-text models show a broad coverage of visual concepts, we find in our experiments that they usually lack the strong discriminative ability required by transfer learning. A natural question is: \emph{can we have one model for both discriminative representations and broad visual concept coverage?}


In this work, we take the first step to answer this question. 
We start with a new perspective, illustrated in Fig.~\ref{fig:teaser}. 
Instead of isolating image-label and image-text data, we define an image-text-label space and show how we can eliminate the boundary between two data types. As shown in Fig.~\ref{fig:teaser} left part, supervised learning~~\cite{khosla2020supervised} on image-label data typically aims at mapping images to discrete labels, and completely ignores the textual concept associated with each label during the training. 
In contrast, language-image contrastive learning~\cite{radford2021learning} aims at learning a pair of visual and textual encoders to align images and texts as shown in Fig.~\ref{fig:teaser} right part. This learning method 
implicitly assumes that each image-text pair has a unique label. Comparing these two learning paradigms side by side, we can see that both of them actually reside in the common image-text-label space, which is constructed by mapping each label to a textual concept for supervised learning, and assigning each textual description a unique label for language-image pretraining, as shown in Fig.~\ref{fig:teaser} bottom. 
Based on this new perspective, we can simply use a visual encoder and a language encoder to encode the images and texts, and align the visual and textual features with the guide of labels (unique labels for image-text pairs and manual labels for image-label data). However, learning from these combined labels cannot be supported in existing supervised learning and language-image contrastive learning paradigms. 
For this purpose, we propose a unified contrastive learning method, called \oursloss{} to seamlessly accommodate both data types for visual-semantic representation learning. It takes images, texts as input and compute the loss with \emph{softened targets} derived from the labels. With \oursloss{}, we combine image-label and image-text data together to learn discriminative \emph{and} semantic-rich representations, which are beneficial to a variety of downstream tasks. To summarize, our main contributions are:
\begin{itemize}[noitemsep,topsep=0pt,parsep=0pt,partopsep=0pt,leftmargin=*]
    \item We introduce a new perspective of image-text-label space, which can seamlessly unify the commonly used image-label and image-text data.     
    \item We propose a unified contrastive learning method called \oursloss{} in the image-text-label space, that can learn from either of the image-label and image-text data, or both.
    \item Extensive experiments show that our \oursloss{} can leverage both types of data effectively and achieve superior performance universally on standard zero-shot, linear probe, fully-finetuning and transfer learning settings.
\end{itemize}

Finally, we scaled up UniCL to billions of image-text-label data in Florence~\cite{yuan2021florence} and demonstrated its superiority over CLIP~\cite{radford2021learning} and ALIGN~\cite{jia2021scaling} across dozens of benchmarks. Hereby, we highly recommend UniCL as a generic multi-modal learning paradigm for vision.

\section{Related works}

\xhdr{Supervised Learning}. Supervised learning for image classification has a long history. As mentioned earlier, a canonical way of supervised learning is mapping images to manual labels. With this goal, numerous works have pushed the image recognition performance from different directions, such as data scale from MNIST~\cite{lecun1989handwritten} to ImageNet-1K~\cite{deng2009imagenet}, model architectures from convolutional neural networks (CNNs)~\cite{lecun1995convolutional,krizhevsky2012imagenet,lin2013network,szegedy2015going,he2016deep,hu2018squeeze,szegedy2016rethinking} to Transformers~\cite{dosovitskiy2020image,wang2021pyramid,liu2021Swin,touvron2021training,wu2021cvt,zhang2021multi,yang2021focal}, and learning objectives from original Cross-Entropy~\cite{murphy2012machine} to marginal losses~\cite{liu2016large,deng2017marginal,sohn2016improved} and recent supervised contrastive loss~\cite{khosla2020supervised}. In this paper, we develop a unified contrastive learning method that regards image-label as image-text-label data to learn a generic visual-semantic space. It calls back the textual concepts behind the labels and use them as a special format of language. In this sense, our work is also related to conventional zero-shot classification~\cite{chollet2016information,xian2016latent,mensink2014costa,wang2018zero,jayaraman2014zero,xian2017zero}. Most of these works pay attention to recognize fine-grained categories at a small scale. Our work goes beyond such restricted setting and is targeted to learn a good and rich visual-semantic representation from the combined image-label and image-text pairs.

\xhdr{Language-Image Contrastive Learning}. Vision-and-language is a rapidly growing field. Existing works can be broadly categorized into two classes. $(i)$ Inspired by the success of BERT~\cite{devlin2018bert}, the first line of research focuses on learning generic multi-modal fusion layers based on masked token prediction and/or image-text matching, given the pre-extracted features from visual and textual encoder~\cite{gordo2017beyond,lu2019vilbert,li2020oscar,su2019vl,zhang2021vinvl,kim2021vilt,li2021align,wang2021simvlm}. They aim to improve downstream tasks such as visual question answering~\cite{antol2015vqa,hudson2019gqa}, image captioning~\cite{lin2014microsoft,agrawal2019nocaps}, visual commonsense reasoning~\cite{zellers2019recognition}.
$(ii)$ 
Another line of works focuses on learning transferable visual representation from natural language supervisions, including generative~\cite{sariyildiz2020learning,desai2021virtex} and contrastive methods~\cite{frome2013devise,wang2016learning, wang2018learning,zhang2020contrastive,radford2021learning,jia2021scaling}. Recently, contrastive learning has been scaled up in representative works such as CLIP~\cite{radford2021learning} and ALIGN~\cite{jia2021scaling}, by pretraining on hundreds of millions of webly-crawled image-text pairs. Our work is close to these works in that we also use the image-text data as one of the major data sources. However, image-label data is ignored in these works. Our work presents the first unified contrastive learning method that can seamlessly leverage both.

\xhdr{Self-supervised Learning}. 
Self-supervised learning (SSL) for vision aims to learn general-purpose visual representations from raw pixels without supervisions from label or text~\cite{goyal2019scaling}. Contrastive learning has laid the foundation for the best performing SSL models~\cite{tian2019contrastive,henaff2019data,chen2020simple,he2020momentum,caron2020unsupervised,tian2020makes,chen2021empirical}. It maximizes agreement of learned representations between differently augmented views of the same image, and minimizes agreement of views from different images. This augmented-view-based paradigm has also been extended to non-contrastive methods~\cite{grill2020bootstrap,chen2021exploring,caron2021emerging,li2021efficient}, where only positive image view pairs are considered in learning. Though image SSL has great promises in leveraging nearly infinite amounts of unlabelled image data in training~\cite{goyal2021self}, the lack of language association renders it hardly applicable to zero-shot recognition. Nevertheless, the success of contrastive learning in SSL has inspired the generalization of this methodology to a much broader range, such as CLIP~\cite{radford2021learning} in image-text setting and our \oursloss{} in image-text-label setting, where images and language descriptions can be considered as multi-modal views of the same underlying concepts.

\section{Method}


\subsection{Preliminaries}
\begin{figure}[t]
	\centering
	\includegraphics[width=0.99\linewidth]{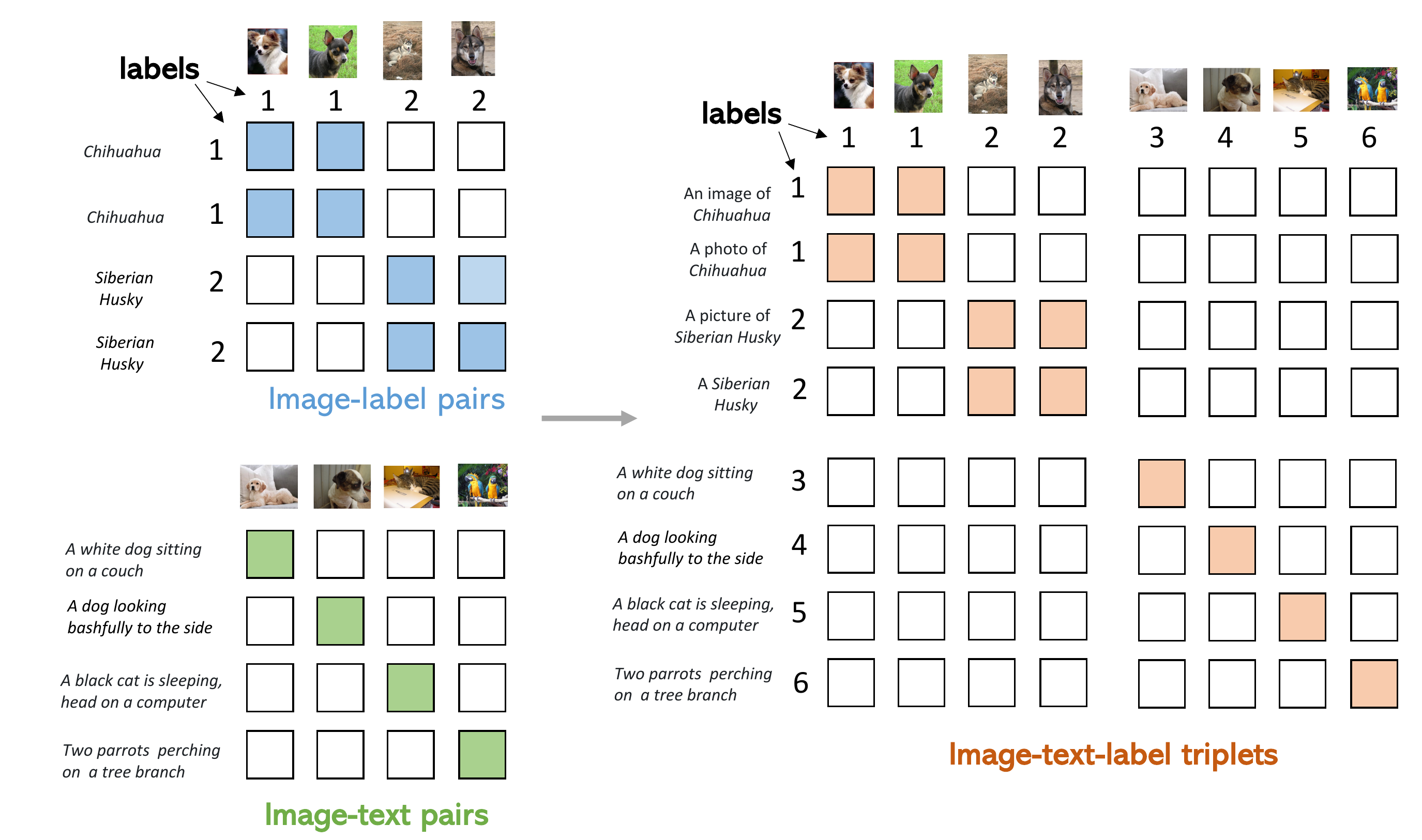}
	\vspace{-6pt}
    \caption{An illustration of covering image-label and image-text data in the image-text-label space. For image-label data, we associate a textual concept to each label, and the images and textual concepts are matched based on the annotated labels (blue tiles). For image-text data, each pair have unique label index, thus matched only at the diagonal entries (green tiles). On the right side, we can simply combine them as image-text-label triplets, and the red tiles means positive pairs while the blank tiles are negative pairs.}
    \label{fig:image-text-label}
    \vspace{-3mm}
\end{figure}

\paragraph{Problem setup.}
We define a triplet-wise data format $\Scal  = \{ (\xv_n, \tv_n, y_n) \}_{n=1}^N$, where $\xv \in \Xcal$  is the image, and $\tv \in \Tcal$ is its corresponding language description (ranging from simple tokens such as category names to free-form text sequences), and $y \in \Ycal$ is a label indicating the index of the grouped or unique language description in the dataset. As we discussed earlier, this triplet data representation is a general format of widely existing image data, including the commonly used image-text and image-label data. On one hand, image-text pairs $\{ (\xv_n, \tv_n) \}_{n=1}^N$ from the web usually have an one-to-one mapping, thus each image-text pair has unique label and $\Scal$ reduces to $\{ (\xv_n, \tv_n, y_n \equiv n) \}_{n=1}^N$. On the other hand, though an image classification problem often uses simple category labels or indices, each label is induced from the similarity of concepts in its task definition~\cite{deng2009imagenet}. Therefore, for image-label data, $\Scal$ reduces to $\{ (\xv_n, \tv_n \equiv C[y_n], y_n) \}_{n=1}^N$, with $C$ as the set of concept names indexed by $y_n$. 
Based on this definition, we can represent an image-label pair as a labeled image-text pair, while an image-text pair as ones with unique label. An example of how they are unified is illustrated in Fig.~\ref{fig:image-text-label}. The goal of this work is to learn from the joint data $\Scal$, believing that the rich semantics in language description $\tv$ and structured organizations of labels $y$ together are beneficial for learning semantic-rich and discriminative visual representations of images $\xv$. 

\subsection{Unified Image-Text-Label Contrast}
\vspace{-2pt}


\begin{figure*}[t!]
	\vspace{-0mm}\centering
\includegraphics[height=4.7cm]{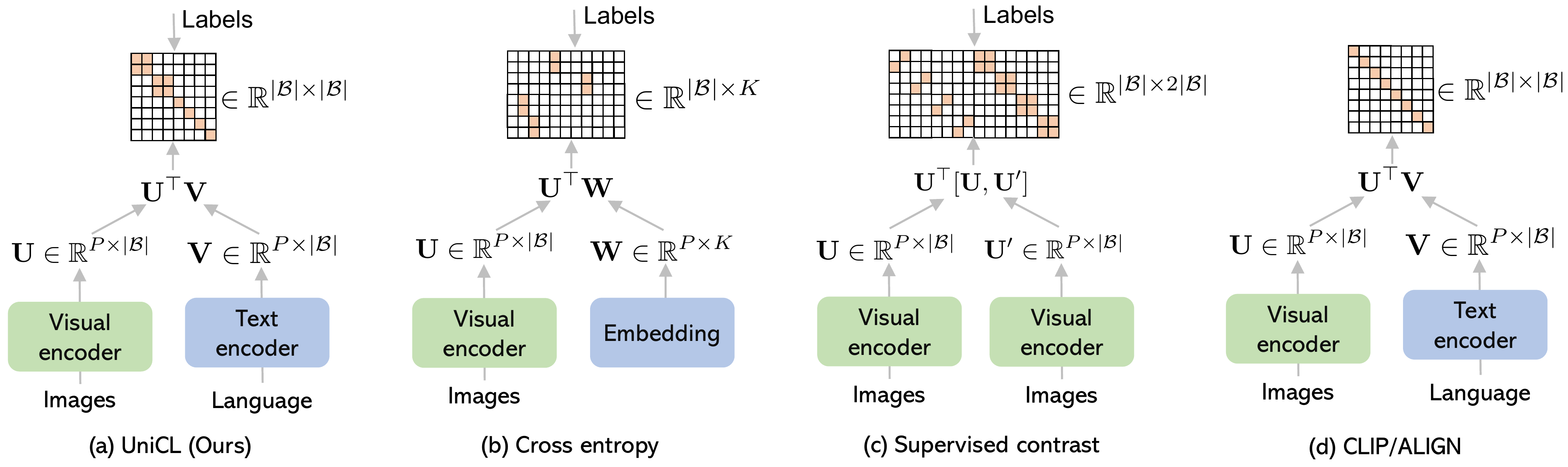} 
	\vspace{-2mm}
	\caption{Illustrative comparisons across different learning paradigms. For a batch of size $|\Bcal|$, all image features $\Umat$, $\Umat^{\prime}$ and text features $\Vmat$ are in dimension $P$, and $K$ is the number of classes. Given a similarity matrix in each method, the labels play the role of defining the positive pairs whose elements are in {\bf \textcolor{orange}{orange}}, negatives are in white; CLIP has the one-to-one assumption for an image-text pair, which implicitly define the diagonal elements as positives.
	 }
	\vspace{-3mm}
	\label{fig:method_comparison}
\end{figure*}


For each image $\xv$, an image encoder model $f_{\thetav}$ parameterized by $\thetav$ first represents $\xv$ as a visual feature vector $ \Tilde{\vv}  \in \R^{d\times 1}$: $ \Tilde{\vv} = f_{\thetav}(\xv)$. For each language description $\tv \in \Tcal$, we encode it with a text encoder $f_{\phiv}(\tv)$ parameterized by $\phiv$ to get its feature vector $ \Tilde{\uv}  \in \R^{d \times 1}: \Tilde{\uv}  = f_{\phiv}(\tv)$. 
For $i$-th image $\xv_i$ and $j$-th language description $\tv_j$ in a batch $\Bcal$, we normalize their feature vector to a hyper-sphere using $ \uv_i = \frac{    f_{\thetav}(\xv_i)  }{   \| f_{\thetav}(\xv_i)  \|} $ and $ \vv_j = \frac{  f_{\phiv}(\tv_j)   }{ \| f_{\phiv}(\tv_j)\| }  $, and their similarity is calculated as $s_{ij}  = \uv_i^{T} \vv_j  $. We consider a bidirectional learning objective between images and language:
\begin{align}\label{eq:obj_bicon}
	\min_{ \{ \thetav, \phiv \} } ~~ \Lcal_{BiC} 	= & \Lcal_{i2t} + \Lcal_{t2i}, 
\end{align}
including two contrastive terms (A temperature hyper-parameter $\tau$ controls the strength of penalties on hard negative samples):

\begin{itemize}[leftmargin=-0.25mm]
\item 
The image-to-text contrastive loss to align matched images in a batch with a given text
\begin{align}\label{eq:obj_i2t_label}
\small
\vspace{-4mm}
\Lcal_{i2t}	= & - \sum_{ i \in \Bcal } \frac{1}{ |\Pcal(i)|  }  \sum_{ k \in \Pcal(i) }
\log \frac{ \exp(\tau \uv_{i}^T \vv_k)  }{\sum_{ j \in \Bcal}  \exp(\tau \uv_{i}^T \vv_{j})  }
\end{align}
where $k \in \Pcal(i) = \{ k | k \in \Bcal, y_k = y_i\}$. 	
\item 
The text-to-image contrastive loss to align matched texts to a given image
\begin{align}\label{eq:obj_t2i_label}
\small
\vspace{-4mm}
\Lcal_{t2i}	= & - \sum_{ j \in \Bcal } \frac{1}{ |\Pcal(j)|  }  \sum_{ k \in \Pcal(j) }
\log \frac{ \exp(\tau \uv_{k}^T \vv_j )  }{\sum_{ i \in \Bcal}  \exp(\tau \uv_{i}^T \vv_{j} )  }
\end{align}
where $k \in \Pcal(j) = \{ k | k \in \Bcal, y_k = y_j\} $.

\end{itemize}





Using Fig.~\ref{fig:image-text-label} right side as an example, the $\Lcal_{i2t}$ is computed for each row, and $\Lcal_{t2i}$ computed for each column. The red tiles indicate the positive pairs while blank tiles the negative ones, all allocated based on the labels.




\subsection{Discussions \& Properties}
\label{sec:connections}
We discuss the unique properties of our proposed \oursloss~and build the connections with previous commonly used learning paradigms. An illustrative comparison is shown in Fig.~\ref{fig:method_comparison}, with more detailed analysis below.

\xhdr{Connections to Cross-Entropy}~\cite{murphy2012machine} We note the proposed  $\Lcal_{BiC}$  in  \eqref{eq:obj_bicon} is closely related to the standard cross-entropy loss used in supervised image classification. Specifically, the text-to-image contrastive term in~\eqref{eq:obj_t2i_label} recovers cross-entropy as a special case, when the following conditions are satisfied:  
$(i)$ the text encoder $f_{\phiv}$ is represented as a simple linear embedding layer $\Wmat$ with a bias $b$. $(ii)$ The batch size $|\Bcal|$ is sufficiently larger than the number of classes $K$, so that all the class embedding vectors are used in contrastive learning, when stochastic sampling is used for training. $(iii)$ $\tau=1$, and $\ell_2$ normalization is excluded, so that  $\Tilde{\uv} = \uv$ and   $\Tilde{\vv}=\vv$. In this case, Eq.~\eqref{eq:obj_t2i_label} becomes: 
\begin{align}\label{eq:objective_ce}
\small
\min_{ \{ \thetav, \Wmat \} } ~~ \Lcal_{CE} =   \sum_{j \in \Bcal}
\log \frac{ \exp(\wv_{\hat{y}}  \Tilde{\vv}_j + b_{\hat{y}} )  }{\sum_{k=1}^K \exp(\wv_{k} \Tilde{\vv}_j +b_{k})  }
\end{align}
where $\hat{y}$ is the ground-truth label for the $j$-th image in the batch.
Based on this, we argue that  $\Lcal_{BiC}$ is more general than  $\Lcal_{CE}$, from two aspects: $(i)$ Augmentation with $\Lcal_{i2t}$. The additional text-to-image term $\Lcal_{i2t}$ in  $\Lcal_{BiC}$ plays the role of regularizer. Given a language description $\tv_j$, all image features with the same $\tv_j$ in the batch are clustered towards the text feature; otherwise they are pushed away. This can help prevent over-fitting, as demonstrated in our experiment later; $(ii)$ Text encoder $f_{\phiv}$. The text encoder can be specified as in more powerful forms such as 12-layer Transformers or pretrained BERT encoder, and take free-form text inputs beyond the set of category names. 

\xhdr{Connections to SupCon~\cite{khosla2020supervised}}
One shared property between our \oursloss{} and SupCon is that both methods exploit label-guided contrastive learning: For any query, both methods leverage samples with the same label to contribute to the numerator as positives. Note that SupCon is proposed in the image-label setting, where each image is augmented with two different views. \oursloss{} and SupCon differ in two aspects: 
$(i)$ {\em Query-vs-Key modality}. 
In SupCon, both query and key in contrastive learning are from the same modality: image-and-image pairs; In \oursloss{}, the query and key are different modalities: image-and-language pairs.
$(ii)$ {\em Encoders.} Only one shared image encoder is used in SupCon for query and key. Two different encoders are used in \oursloss{} for different modalities, as shown in Fig.~\ref{fig:method_comparison}.

\xhdr{Connections to CLIP~\cite{radford2021learning}}
For image-texts pairs, there are only one-to-one mappings between an image and its paired text in a batch. In another word, $ \Pcal(i) = \{i\}$ and $\Pcal(j) = \{j\}$ for Eq.~\eqref{eq:obj_i2t_label} and Eq.~\eqref{eq:obj_t2i_label}, respectively. Then $\Lcal_{BiC}$ becomes:
 
\vspace{1mm}
\begin{minipage}{0.46\textwidth}
\centering

\begin{itemize}[leftmargin=0.5mm]
\item 
The image-to-text contrastive loss
\begin{align}\label{eq:obj_i2t_caption}
\small
\Lcal_{i2t}	= & - \sum_{ i \in \Bcal } 
\log \frac{ \exp(\tau \uv_{i} \vv_i )  }{\sum_{ j \in \Bcal}  \exp(\tau \uv_{i} \vv_{j} )  }
\end{align}

\item 
The text-to-image contrastive loss
\begin{align}\label{eq:obj_t2i_caption}
\small
\Lcal_{t2i}	= & - \sum_{ j \in \Bcal } 
\log \frac{ \exp(\tau \uv_{j} \vv_j )  }{\sum_{ i \in \Bcal}  \exp(\tau \uv_{i} \vv_{j} )  }
\end{align}

\end{itemize}
\vspace{1mm}
\end{minipage}

This means that $\Lcal_{BiC}$ reduces to CLIP training objective, when only image-text data is employed. The major structural change of \eqref{eq:obj_i2t_label} over \eqref{eq:obj_i2t_caption} is that for each language description, any of the image samples with the same label are considered as positives in a batch, contributing to the numerator. Similar conclusion is drawn by comparing \eqref{eq:obj_t2i_label} and \eqref{eq:obj_t2i_caption}.

\begin{algorithm}[!ht]
  \caption{Training process for \oursloss.}
  \label{alg:latex}
  \footnotesize
  \Comment{n: batch size; d: projected feature dim}
  \Comment{The main training loop}
    \For{ \var{$\xv$}, \var{$\tv$}, \var{y}  in \var{loader}}{
    \var{target} = \FuncCall{TargetM}{\var{y}} \;
    \Comment{ Image encoding: \!\!n$\times$d } 
    \var{u} =  \FuncCall{l2\_normalize}{\FuncCall{$f_{\thetav}$}{\var{$\xv$}}, dim=-1}\; 
    \Comment{ Text encoding: \!\!n$\times$d} 
    \var{v} = \FuncCall{l2\_normalize}{\FuncCall{$f_{\phiv}$}{\var{$\tv$}}, dim=-1}\;  
     
    \Comment{ Cosine similarities:  \!\!n$\times$n } 
    \var{logits = \var{exp}($\tau$) $\cdot$  u * v.T  }   
   
   \Comment{ Bidirectional contrastive loss } 
   \var{i2t} = \FuncCall{SoftCE}{logits, target} \;
   \var{t2i} = \FuncCall{SoftCE}{logits.T, target.T} \;
   \var{loss = (\var{i2t} + \var{t2i})/2 } \;
   \var{loss.backward() } \;
     
    }
  \Comment{The Target Modification function}
  \Function{TargetM(\var{y})}{
    \Comment{Note y = 0  for image-text in loader}
     \var{cap\_m  = (\var{y} == 0).sum() }\;
     \var{cls\_m  = \var{y}[\var{y} > 0].max()  }\;
     \var{y[y == 0] =    arange(0, cap\_m) + cls\_m + 1} \;
    \Return {  y.view(-1, 1) == y.view(1, -1) } \;
  }        
  \Comment{The SoftTargetCrossEntropy function}
  \Function{SoftCE(\var{s}, \var{t})}{
     \var{s  = softmax(\var{s}, dim=-1)} \;
     \var{loss = - (t * log(s)).sum(dim=-1)} \;
    \Return { (loss/t.sum(dim=-1)).mean() } \;
  }    
\end{algorithm}

\subsection{Model Training and Adaptation}

The training process of \oursloss~is summarized in Algorithm~\ref{alg:latex}. Note that this pseudo code is related to our data loader construction: all the image-text pairs have an initial label index $y=0$, while all image-label pairs have an initial label index $y \in [1, \cdots, K]$. The $\mathbf{TargetM}$ function ensures that each unique language description in the batch has a unique label index. In training, $\tau$ is a trainable variable initialized as 1.
%
After training, the learned visual and textual encoder $\{f_{\thetav}, f_{\phiv}\}$ can be used jointly for open-vocabulary image recognition, \textit{i.e.}, recognizing the categories seen during training or novel ones beyond the annotated categories. Alternatively, the visual backbone $f_{\thetav}$ can be used independently, either for feature extraction in linear probe or for full model finetuning in object detection.

\section{Experiments}

In this section, we examine \oursloss{} to answer two research questions. 
\textbf{\texttt{Q1}} learning objective -- {how does our \oursloss{} perform compared with CE and SupCon on image classification?} 
\textbf{\texttt{Q2}} pre-training data -- {what is the unique benefit of applying \oursloss{} on the joint image-text-label data?} 

\xhdr{Datasets}. We study our models based on publicly available datasets, and the statistics are shown in Table~\ref{tab:dataset}. For classification data (top four rows), the number of visual concepts are identical to the number of categories. For image-text data (bottom three rows), we use Spacy~\cite{spacy2} to extract the noun phrases and then count the number of unique noun entities that appear more than 5 times. Given the pool of concepts, we then calculate the number of unique words and report it as the vocabulary size.
The ratio of \#images/\#concepts clearly illustrates the different trade-off between image diversity and semantic-richness over different datasets. In our unified image-text-label space, all these datasets are homogeneous, and can be jointly used for learning. GCC-15M denotes the merged version of GCC-3M and 12M.

\xhdr{Training}. We use the same prompt strategy and tokenizer for classification data as proposed in CLIP~\cite{radford2021learning}. We fill the class names into the prompt templates, followed by a tokenization before feeding into the text encoder. During training, we randomly sample one prompt template while averaging over all 80 templates for validation. For fair comparison, we use the same text encoder architecture as in CLIP~\cite{radford2021learning}, and the whole model including vision and text encoder are trained from scratch. More training details are discussed in the following individual sections.

\xhdr{Evaluation}. We evaluate the quality of learned representations on a set of computer vision tasks, including:

\begin{minipage}{0.46\textwidth}
\centering
\hspace{-4mm}
\begin{itemize}[leftmargin=.5mm]
\item {\em Standard classification}. We report the Top-1 classification accuracy on CIFAR-10~\cite{krizhevsky2009learning}, CIFAR-100~\cite{krizhevsky2009learning} and ImageNet-1K~\cite{deng2009imagenet}.
\end{itemize}
\end{minipage}

\begin{minipage}{0.46\textwidth}
\centering
\hspace{-4mm}
\begin{itemize}[leftmargin=.5mm]
\item {\em Zero-shot classification}. We evaluate on ImageNet-1K as well as 14 datasets used in~\cite{radford2021learning}, and employ the same text prompts. Averaged $\texttt{scores}$ is reported.
\end{itemize}
\end{minipage}

\begin{minipage}{0.46\textwidth}
\hspace{-4mm}
\centering
\begin{itemize}[leftmargin=.5mm]
\item {\em Linear probe}. 
We study 18 datasets used in~\cite{radford2021learning}. Automatic hyper-parameter tuning is considered to ensure fairness of comparison. The averaged $\texttt{scores}$ is reported.
\end{itemize}

\end{minipage}

\begin{minipage}{0.46\textwidth}
\hspace{-2mm}
\centering
\begin{itemize}[leftmargin=.5mm]
\item {\em Object detection}. We use Mask R-CNN~\cite{he2017mask} as the detector and follow the standard 1$\times$ schedule. $\texttt{mAP}$ for box and mask are reported on 80 object categories.
\end{itemize}

\end{minipage}

\begin{table}[t]
    \centering
    \footnotesize
\setlength{\tabcolsep}{2.5pt}
    \begin{tabular}{lcccc}
    \toprule
    Dataset & \#Images & \#Concepts & Vocab. Size & \#Img/C.  \\
    \midrule
    CIFAR-10~\cite{krizhevsky2009learning} & 50k & 10 &  10 & 5000 \\
    CIFAR-100~\cite{krizhevsky2009learning} & 50k & 100 & 105 & 500 \\
    ImageNet-1K~\cite{deng2009imagenet} & 1.3M & 1,000 & 1,233 & 1300 \\
    ImageNet-22k~\cite{deng2009imagenet} & 14.2M & 21,841 & 14,733 & 650 \\
    GCC-3M~\cite{sharma2018conceptual} & 3.3M & 17,135 & 7,953 & 193 \\
    GCC-12M~\cite{changpinyo2021conceptual} & 12M & 584,261 & 98,347 & 21 \\
    YFCC-14M~\cite{thomee2016yfcc100m} & 14M & 650,236 & 214,380 & 22 \\
    \bottomrule
    \end{tabular}
    \vspace{-7pt}
    \caption{Statistics of training datasets used in our experiments. \#Img/C. is ratio between the numbers of images and concepts.}
    \label{tab:dataset}
    \vspace{-5mm}
\end{table}

\begin{table*}[t]
\centering
\footnotesize
\setlength{\tabcolsep}{2.5pt}
    \begin{tabular}{lcc|cc|cccc}
    \toprule
    \multirow{2}{*}{Method} & \multicolumn{2}{c|}{CIFAR-10} & \multicolumn{2}{c|}{CIFAR-100} & \multicolumn{4}{c}{ImageNet-1K}
    \\
    \cmidrule{2-9}
           & ResNet-50 & ResNet-101 & ResNet-50 & ResNet-101 & ResNet-50 & ResNet-101 & Swin-Tiny & Swin-Tiny$^\dagger$\\
    \midrule
    CrossEntropy & 95.0 & 96.5 & 75.3 & 78.8 &78.2  & 79.8 & 76.8  & 81.4 \\
    SupCon~\cite{khosla2020supervised} & 96.0 & 96.8 & 76.5 & 79.6 & \textbf{78.7} & \textbf{80.2} & 77.0 & n/a \\
    \oursloss~(Ours) & \textbf{96.8} & \textbf{97.0} & \textbf{78.4} & \textbf{81.4} & 78.1 & 79.9 & \textbf{79.9} & \textbf{81.7} \\
    \bottomrule
    \end{tabular}
    \vspace{-3mm}
    \caption{Image classification trained with CE, SupCon~\cite{khosla2020supervised} and our \ourslossfull. ResNet-50~\cite{he2016deep}, ResNet-101~\cite{he2016deep} and Swin Transformer Tiny~\cite{liu2021Swin} are used as the visual encoders. $\dagger$ means trained with MixUp~\cite{zhang2017mixup} and CutMix~\cite{yun2019cutmix} data augmentation as in~\cite{touvron2021training}. The following numbers are from~\cite{khosla2020supervised}: ResNet-50 trained on CIFAR-10 and CIFAR-100, ResNet-50 and ResNet-101 on ImageNet-1K. Since there is no clear way to use CutMix or MixUp in SupCon, we leave it as ``n/a'' for Swin-Tiny$^\dagger$ model. }
    \label{tab:image_recognition}     
    \vspace{-3mm}
\end{table*}

\begin{table}[]
    \centering
    \footnotesize
    \begin{tabular}{ccl|c}
    \toprule
          $\Lcal_{i2t}$ & $\Lcal_{t2i}$ & {Text Encoder} & {Top-1 Acc.} \\
          \midrule
        \checkmark & \checkmark &
          Transformer & {79.9} \\
          \checkmark & \checkmark & Embedding & 78.7 \\
          - & \checkmark & Embedding & 75.7 \\ 
         \bottomrule
    \end{tabular}
    \vspace{-3mm}
    \caption{Performance with different losses and text encoders.}
    \label{tab:perf_w_loss_textencoder}
    \vspace{-3mm}
\end{table}

\begin{table}[]
    \centering
    \footnotesize
    \begin{tabular}{lccc|ccc}
    \toprule
     & \multicolumn{3}{c|}{Transformer Encoder} & \multicolumn{3}{c}{Simple Embedding} \\
     \cmidrule{2-7}
          Batch Size & 1024 & 2048 & 4096 & 1024 & 2048 & 4096 \\
          \midrule
          Top-1 Acc. & 79.9 & 80.1 & 79.9 & 79.0 & 78.9 & 78.7  \\
         \bottomrule
    \end{tabular}
    \vspace{-3mm}
    \caption{Performance of \oursloss~with respect to different batch sizes. The number of training epochs is kept the same.}
    \vspace{-3mm}
    \label{tab:perf_with_bs}
\end{table}

\subsection{Results of \oursloss{} on image classification}
\label{subsec:learn_w_image_label}

To gain empirical understanding of our \oursloss{} objective, we compare \oursloss{} against two supervised learning methods, Cross-Entropy (CE)~\cite{murphy2012machine} and Supervised Contrastive Learning (SupCon)~\cite{khosla2020supervised} on image classification datasets. We employ two representative architectures,  ResNet~\cite{he2016deep} and Swin Transformer~\cite{liu2021Swin} to build the visual encoder, whose last layer output are pooled as the visual representation. We use standard random crop as the data augmentation. All models are trained for 500 epochs with a batch size of 4096. 
%
We report the comparison results in Table~\ref{tab:image_recognition}, Overall, the proposed \oursloss~achieves comparable if not better performance across all datasets and model architectures. 

\xhdr{Comparison with SupCon}~\cite{khosla2020supervised}. We can find that our \oursloss{} is superior on CIFAR-10 and CIFAR-100 and on par with SupCon on ImageNet-1K. Both \oursloss~and SupCon pursue bidirectional alignments, one for image-text pairs and the other one for images from multi-views. Though the overall performance is comparable on these standard classification tasks, our \oursloss~has two unique advantages over SupCon: 1) it is end-to-end training while SupCon requires two training stages, \textit{i.e.}, visual encoder training and a linear classifier tuning; 2) the learned representations in our model is language-aware, which means we can directly use it for zero-shot recognition, as demonstrated later.

\xhdr{Comparsion with CE}~\cite{murphy2012machine}. \oursloss{} in~\eqref{eq:obj_bicon} promotes a bidirectional alignment between images and category names, which imposes an additional regularization term than CE in~\eqref{eq:objective_ce}.  
As such, it can be particularly helpful when over-fitting tends to occur. For example, when training ResNet-50 on small datasets such as CIFAR-10 and CIFAR-100, \oursloss{} improves around 1-3 points over CE. 
When training Swin Transformer on ImageNet-1K, the network tends to over-fit due to the lack of spatial inductive bias; Our \oursloss{} outperforms CE by 3 points. 
When over-fitting is less severe, such as training on larger datasets (from CIFAR to ImageNet) or with strong augmentation (MixUp~\cite{zhang2017mixup} and CutMix~\cite{yun2019cutmix}), our method is still on par with CE.

\begin{table*}[!ht]
    \centering
    \footnotesize
    \begin{tabular}{lc|cccccc}
    \toprule
        \multirow{3}{*}{Training Data} &  \multirow{3}{*}{Method} &  \multicolumn{4}{c}{Metric}  \\
        \cmidrule{3-7}
        & & \multirow{2}{*}{ImageNet-1K} & \multirow{2}{*}{\makecell{Zero-shot \\ 14 datasets}} & \multirow{2}{*}{\makecell{Linear probe \\ 18 datasets}} & \multicolumn{2}{c}{COCO detection} \\ 
        & & & & & box mAP & mask mAP \\
        \midrule
        ImageNet-1K   & CrossEntropy    & 76.8 & n/a & 78.1 & 42.6 & 39.5 \\
        ImageNet-1K   &  SupCon  & 77.0  & n/a & 70.6 & 42.5 & 39.3    \\
        ImageNet-1K & \oursloss & 79.9 & 30.2 & 78.0 & 42.5 & 39.4 \\
        \midrule 
        ImageNet-1K + GCC-3M  &   \oursloss & 80.2 & 39.0 & 78.9 &  43.0 & 39.5 \\           
        ImageNet-1K + YFCC-14M  & \oursloss & 81.1 & 40.0 & 80.1 & 42.5 & 39.3 \\
        ImageNet-1K + GCC-15M  &  \oursloss & \textbf{81.8} & \textbf{45.1} & \textbf{81.5}  & \textbf{43.7} & \textbf{40.3} \\                 
        \bottomrule
    \end{tabular}
    \vspace{-3mm}
    \caption{Performance for various training objectives and adding image-text pairs to ImageNet-1K dataset.}
    \vspace{-3mm}
    \label{tab:imagenet1K_plus_imagetext}
\end{table*}

\xhdr{Ablation of language encoders $f_{\phiv}$}. Our \oursloss{} has the flexibility in constructing its language encoders. In Table~\ref{tab:perf_w_loss_textencoder}, we ablate by comparing two options: Transformers {\em vs} a simple linear embedding layer $\Wmat$. The former is superior by absolute 1.2\%. We suspect this is due to its ability to capture the semantics behind the 1K category names. For example, two categories ``tree frog'',``tailed frog'' share the common word ``frog'', which conveys a language prior knowledge about their similarity. This semantic information, however, can be hardly captured by an embedding layer indexed with labels. One may notice that our model using extra language encoder introduces more parameters, leading to an unfair comparison. However, during inference, the language encoder is used to extract the textual embeddings for all concepts and then discarded. Therefore, the effective complexity and time cost during inference is nearly identical to the other methods.

\xhdr{Ablation of training objectives}. The third row in Table~\ref{tab:perf_w_loss_textencoder} shows a significant 3\% drop by remaining the term $\Lcal_{t2i}$ only in our bidirectional loss. It indicates the  importance of both loss terms in our \oursloss{}. Though $\Lcal_{t2i}$ resembles CE under certain conditions described in Section~\ref{sec:connections}, we notice a small gap between them (75.7 \textit{v.s.} 76.8 in Table~\ref{tab:image_recognition}). This gap is probably attributed to the stochastic training. At each iteration, CE always compares the visual feature to the entire 1K class embeddings, while the \oursloss{} updates with the subset of concept embeddings in the current mini-batch.

\xhdr{Effect of training batch size}. We vary the default batch size from 4096 to 2048 and 1024.
Results are shown in Table~\ref{tab:perf_with_bs}. \oursloss{} is robust to the variation of batch size, regardless of which language encoder is employed.
This observation is different from contrastive methods such as SimCLR~\cite{chen2020simple} in self-supervised learning. This is probably because: $(i)$ one of the two views is the embeddings of category names in our \oursloss{}, which are consistently used with high overlap across different mini-batches, which make the learning less vulnerable to the batch size; $(ii)$ The label information provides a consistent and strong guidance.

\subsection{Results on data unification of image-text-label }
\label{subsec:leaning-with-image-text-label}
\vspace{-3pt}
In this part, we study the benefits of \oursloss{} when learned with the unification of image-label and image-text data. We use Swin-Tiny as the visual encoder for consistency.

\subsubsection{Benefit of image-text to image-label}
\vspace{-2pt}
\label{subsec: benefit-of-image-text}
We use ImageNet-1K as the base dataset, and gradually add different sets of image-text pairs, including GCC-3M, GCC-15M and YFCC-14M. When combining with image-text pairs, we use a balanced data sampler to ensure that the model is trained with the same number of image-label and image-text pairs per epoch. All models are trained with 500 epochs.  We report the results in Table~\ref{tab:imagenet1K_plus_imagetext}.

\xhdr{Comparison of objectives.} From the first three rows, we see that the models trained with different objectives on ImageNet-1K obtain similar performance across different metrics. However, our \oursloss{} is the only one that is directly applicable for zero-shot image recognition, though CE can be partially used for zero-shot with extra label mapping efforts. Surprisingly, the average zero-shot performance over 14 datasets for \oursloss{} trained only on ImageNet-1K reaches a similar level to CLIP trained on YFCC-14M (30.2 \textit{v.s.} 36.3 as will be shown in Table~\ref{tab:perf_imagelabel_to_imagetext_a}).

\xhdr{Benefit of image-text pairs.} Adding image-text pairs can generally improve the performance across all metrics. In the table, we can see all image-text datasets help to significantly improve the zero-shot performance. Besides, adding GCC-3M further improves linear probe and COCO detection by 0.9 and 0.5, respectively. YFCC-14M helps to improve ImageNet-1K and linear probe by 1.2 and 2.1, respectively. As summarized in Table~\ref{tab:dataset}, image-text pairs are coarsely aligned, but cover rich visual concepts. Therefore, they are particularly beneficial for tasks requiring broad visual concept understanding, such as zero-shot and linear probe on dozens of datasets. 
When GCC-15M is used, we observed much more improvements as well for ImageNet-1K (+1.9), Linear Probe (+3.5) and COCO detection (+1.2). Note that we used balanced data sampler to ensure the model sees equal number of image-text batches during training. This suggests that concept richness (GCC-15M is much higher than GCC-3M) and quality (GCC-15M is much cleaner than YFCC-14M) are both important to compensate classification data for learning discriminative representations.

\begin{figure}[t]
    \centering
    \begin{minipage}{0.48\linewidth}
    \noindent\includegraphics[width=0.99\linewidth]{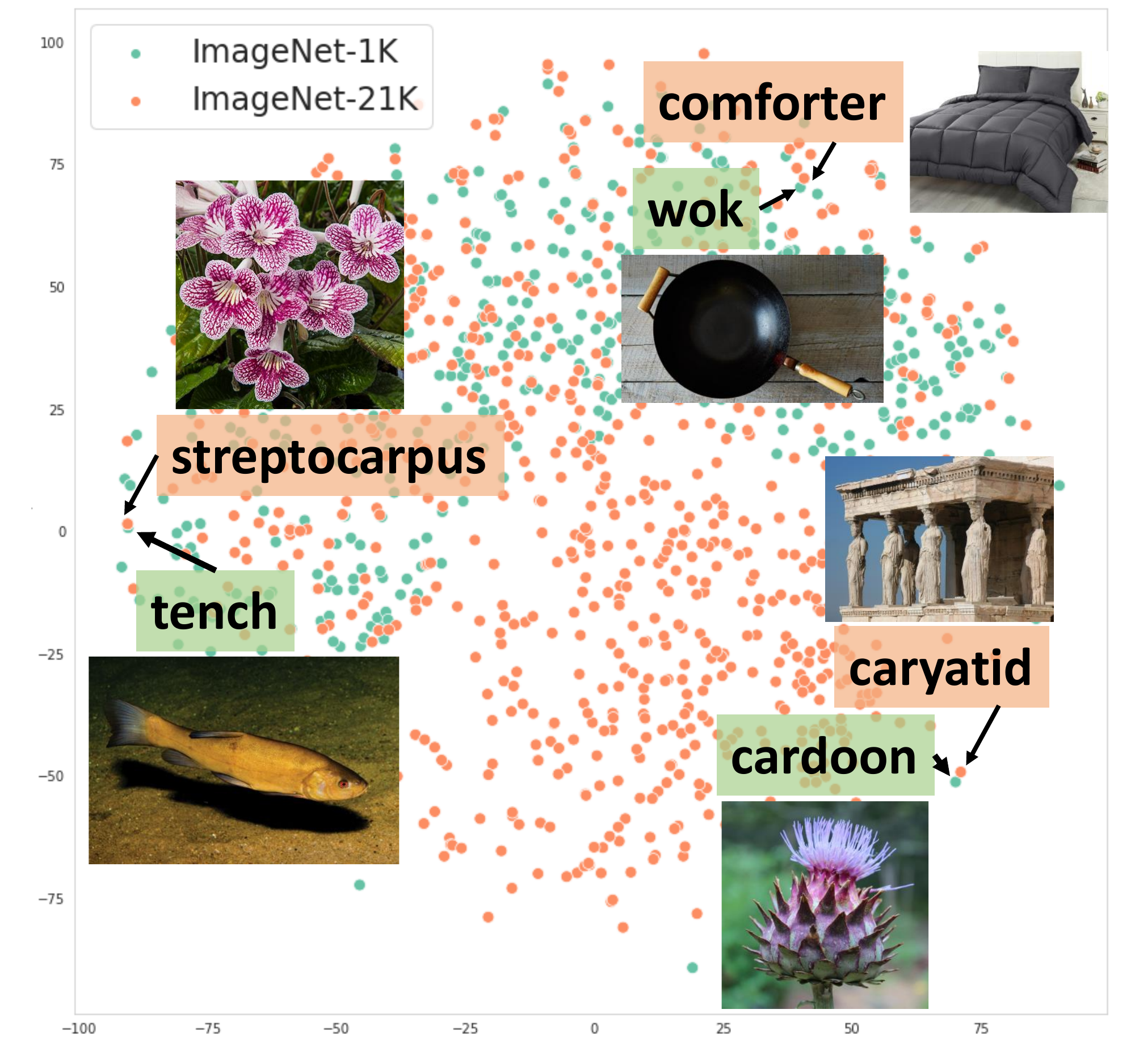}    
    \end{minipage}   
    \begin{minipage}{0.48\linewidth}
    \noindent\includegraphics[width=0.99\linewidth]{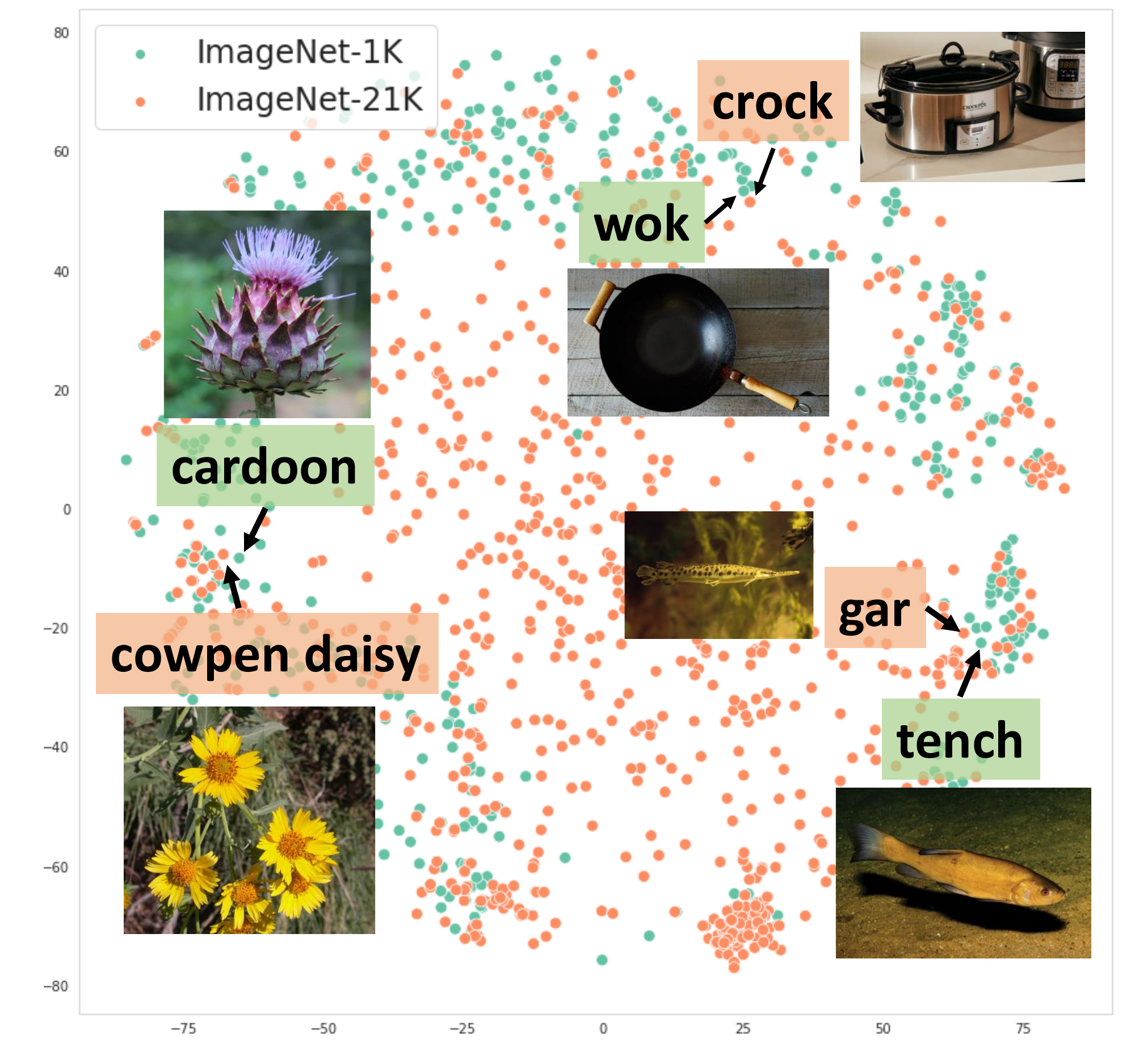}         
    \end{minipage}  
    \vspace{-3mm}
    \caption{2D $t$-SNE visualization of textual concepts encoded by the learned text encoder. We plot 1000 classes for both ImageNet-1K and ImageNet-21K. Given the category from ImageNet-1K (\textcolor{LimeGreen}{\bf green box}), we find the closest category from ImageNet-21K (\textcolor{orange}{\bf orange box}). Left: \oursloss{} trained on ImageNet-1K; Right: \oursloss{} trained on ImageNet-1K+GCC-15M. Better viewed in color.}
    \vspace{-3mm}
    \label{fig:visualize_semantic_space}
\end{figure}

For qualitative analysis, we visualize the 2D $t$-SNE~\cite{van2008visualizing} of the textual feature space in Fig.~\ref{fig:visualize_semantic_space}. Given a query concept from ImageNet-1K, we search the closest target concept from the remained 21K concepts in ImageNet-22K in the feature space. For better understanding, we also show the exemplar image corresponding to each concept. Clearly, model trained on ImageNet-1K can hardly generalize to understand the concepts from the other 21K concepts. In contrast, adding GCC-15M image-text pairs significantly improve the its understanding ability, as the retrieved target become more semantically similar to the queries in ImageNet-1K.

\begin{table*}[t]
    \centering
    \footnotesize
    \begin{tabular}{lc|ccccc}
    \toprule
        \multirow{4}{*}{Training Data} & \multirow{4}{*}{Method} & \multicolumn{4}{c}{Metric}  \\
        \cmidrule{3-6}
        & & \multicolumn{2}{c}{Zero-shot} & \multirow{2}{*}{\makecell{ImageNet-1K \\ Finetuning}}& \multirow{2}{*}{\makecell{Linear Probe \\ 18 datasets}}\\
        \cmidrule{3-4}
        & & ImageNet-1K & 14 datasets  &  &    \\
        \midrule
        YFCC-14M & CLIP & 30.1 & 36.3 & 77.5  &  72.7  \\
        ImageNet-21K & \oursloss & 28.5 &   37.8 & 78.8 & 80.5 \\
        YFCC-14M + ImageNet-21K~(half/half) & Multi-task & 33.0 & 41.5 & 78.0 &  74.1 \\           
        YFCC-14M + ImageNet-21K~(half/half) & \oursloss & 36.4 & 45.5 & 79.0 &  80.0 \\           
        YFCC-14M + ImageNet-21K & \oursloss & \textbf{40.5} & \textbf{49.1} & \textbf{80.2}  & \textbf{81.6} \\        
        \midrule
        ImageNet-22K & \oursloss & 66.8 & 38.9 & 80.3 &  82.0    \\   
        YFCC-14M + ImageNet-22K & Multi-task & 40.9 &  47.6 & {80.4} &   82.0   \\        
        YFCC-14M + ImageNet-22K & \oursloss & 70.5  & 52.4  & \textbf{80.5} &    82.0\\
        GCC-15M \hspace{0.8mm} + ImageNet-22K & Multi-task & {50.6} &  {51.8} & 79.9 &  \textbf{82.5} \\           
        GCC-15M \hspace{0.8mm} + ImageNet-22K & \oursloss & \textbf{71.3} &  \textbf{53.8} & 80.0  & {82.1} \\
        \bottomrule
    \end{tabular}
    \vspace{-3mm}
    \caption{Ablation studies on the training datasets and tasks. Each model is pre-trained with 32 epochs following CLIP~\cite{radford2021learning}.}
    \label{tab:perf_imagelabel_to_imagetext_a}
      \vspace{-3mm}
\end{table*}

\begin{figure*}[t]
    \centering
    \noindent\includegraphics[width=0.93\textwidth]{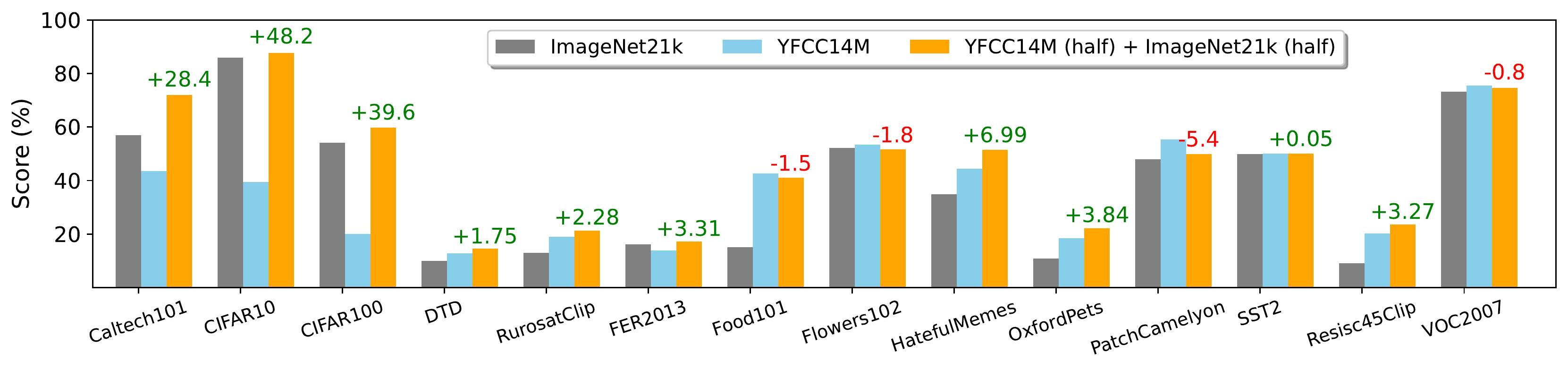}    
    \vspace{-4mm}
    \caption{Zero-shot classification on 14 datasets. The gain between \oursloss{} on mixed data and CLIP on YFCC-14M data is shown. \oursloss{} combines the advantages of learning rich concept coverage from image-text pairs and discriminative representations from image-label data: \oursloss{} outperforms both baselines significantly on the first 3 datasets, and shows higher averaged scores on others.}
    \label{fig:transfer_zs_cls}
    \vspace{-3mm}
\end{figure*}

\subsubsection{Benefit of image-label to image-text}
\vspace{-2pt}
We switch the role to study how image-label data can assist the learning with image-text pairs. Follow the protocols in CLIP~\cite{radford2021learning}, we use random crop as the data augmentation, a standard data sampler, and train all models for 32 epochs. 
We compare against two baselines: 
$(i)$ {\em CLIP}, a language-image contrastive learning method without label supervision, our \oursloss~can recover CLIP when merely using image-text pair for the training. 
$(ii)$ {\em Multi-task} learning that performs CE on image-label data, and CLIP on image-text data.

We report the results in Table~\ref{tab:perf_imagelabel_to_imagetext_a}. We first reproduced CLIP on YFCC-14M with Swin-Tiny. The ImageNet-1K zero-shot accuracy is 30.1\%, which closely matches the reported number 31.2\% with ResNet-50 in~\cite{radford2021learning}. 
To ensure fair comparisons, we build a ImageNet-21K dataset by excluding the categories in ImageNet-1K from ImageNet-22K dataset, and train \oursloss{}. Interestingly, it achieves comparable ImageNet-1K zero-shot performance to YFCC-14M. This indicates image-label data is arguably another good source of learning visual-semantic representations, which is nevertheless less studied in previous works. 
We combine half of ImageNet-21K and YFCC-14M datasets so that the total number of training instances remains the same, and train a \oursloss{} model. This data unification boosts performance almost on all metrics, especially on zero-shot classification for ImageNet-1K (absolute $6\%>$ gain) and 14 datastes (absolute $7\%>$ gain). The detailed comparison on 14 datasets in Fig.~\ref{fig:transfer_zs_cls}, shows that \oursloss{} wins on 11 out of 14 datasets. Besides zero-shot, our \oursloss{} also achieves significant improvement (+{7.3}\%) on linear probe compared with the CLIP baseline. With the full set of both datasets (row 4), the performance can be uniformly improved further. 

\begin{figure}[t]
    \centering
    \begin{minipage}{0.48\linewidth}
    \noindent\includegraphics[width=0.99\linewidth]{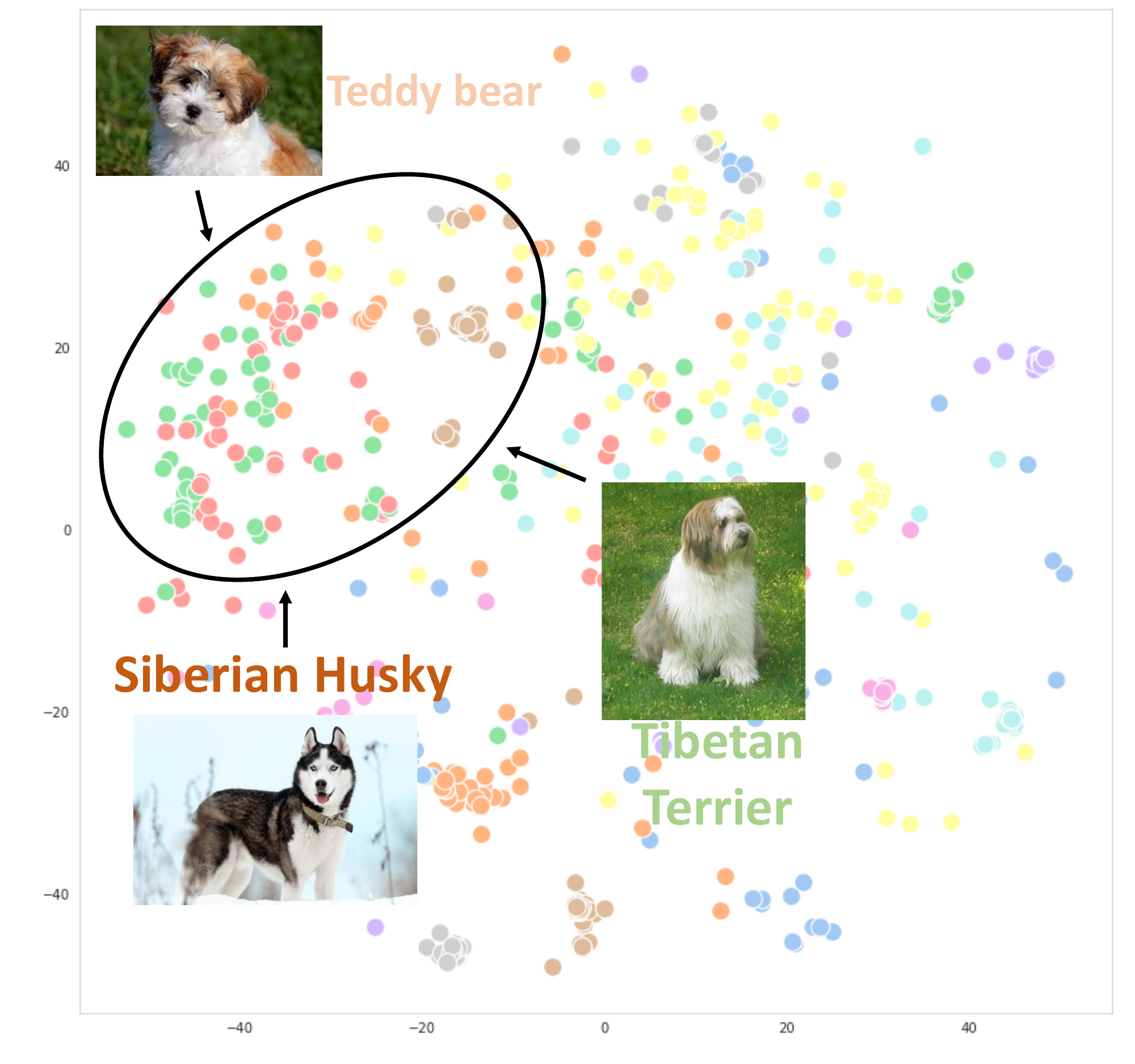}        
    \end{minipage}
    \begin{minipage}{0.48\linewidth}
    \noindent\includegraphics[width=0.99\linewidth]{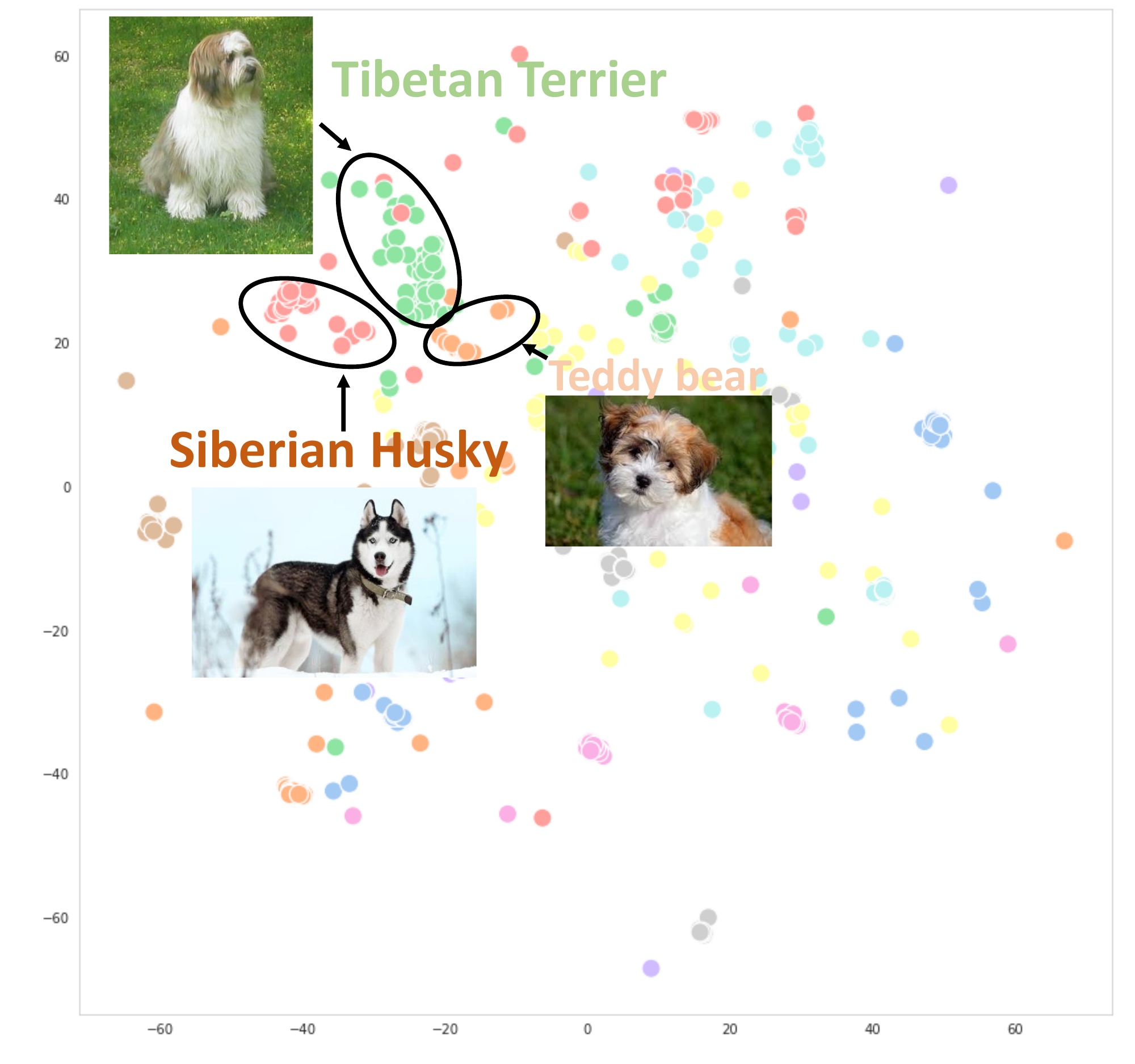}        
    \end{minipage}    
    \vspace{-3mm}
    \caption{2D $t$-SNE visualization of visual features from visual encoders. We randomly select images from 20 classes of ImageNet-1K and visualize the distribution for Left: CLIP trained on YFCC-14M; Right: \oursloss{} trained on YFCC-14M~(half)+ImageNet-21K~(half). Three categories ``teddy bear'',``siberian husky'' and ``tibetan terrier'' are highlighted. Better viewed in color.}
    \vspace{-4mm}
    \label{fig:visualize_visual_space}
\end{figure}

We compare our method with multi-task learner with different datasets. First, when using half of YFCC-14M and ImageNet-21K, our \oursloss{} outperforms multi-task learner by a large margin across all tasks. When trained with the ImageNet-22K, the gaps shrink for ImageNet-1K finetuning and linear probe but remain for zero-shot recognition. This is mainly because ImageNet-22K cover all ImageNet-1K concepts and a large portion of categories in the linear probe datasets. Admittedly, Multi-task learner is a good representation learning method. However, because it isolates image-label and image-text pairs, it cannot learn a discriminative and semantic-rich feature space as our method.

Finally, to qualitatively show that how \oursloss{} trained with image-label data yields a more discriminative feature space, we visualize the 2D $t$-SNE for the visual features of ImageNet-1K dataset in Fig.~\ref{fig:visualize_visual_space}. Dogs with fine-grained breeds are heavily mixed together for the model trained on image-text pairs only. However, they are clearly grouped with the aid of image-label data from ImageNet-21K, even though it contains none of those dog breed concepts.

\section{Conclusion}
\vspace{-2pt}
We have presented \oursloss{}, a new contrastive learning paradigm for generic multi-modal representation learning. It is built in the image-text-label space, and empowered by our unified contrastive learning method. Such a unified paradigm prompts a seamless synergy between image-label and image-text pairs for discriminative and semantic-rich representation learning, which brings universal improvements on zero-shot, linear probe, finetuning benchmarks. 
Moreover, we discuss its connections to existing learning methods, and empirically demonstrated that our learning method stand-alone is a good alternative learner on pure image-label data.

\xhdr{Discussions}. During our submission, we mainly focused on vision tasks such as image recognition and object detection, and based our model on public datasets. However, we refer the readers to Florence~\cite{yuan2021florence} for large-scale pretraining and evaluation on a boarder set of tasks including VQA and video understanding. We note that Florence used a huge amount of private data and thus recommend the suite of experiments in this paper as a baseline for future academic research.



{\small
\bibliographystyle{ieee_fullname}
\bibliography{unicl}

\begin{thebibliography}{10}\itemsep=-1pt

\bibitem{agrawal2019nocaps}
Harsh Agrawal, Karan Desai, Yufei Wang, Xinlei Chen, Rishabh Jain, Mark
  Johnson, Dhruv Batra, Devi Parikh, Stefan Lee, and Peter Anderson.
\newblock nocaps: novel object captioning at scale.
\newblock In {\em Proceedings of the IEEE/CVF International Conference on
  Computer Vision}, pages 8948--8957, 2019.

\bibitem{antol2015vqa}
Stanislaw Antol, Aishwarya Agrawal, Jiasen Lu, Margaret Mitchell, Dhruv Batra,
  C~Lawrence Zitnick, and Devi Parikh.
\newblock Vqa: Visual question answering.
\newblock In {\em Proceedings of the IEEE international conference on computer
  vision}, pages 2425--2433, 2015.

\bibitem{caron2020unsupervised}
Mathilde Caron, Ishan Misra, Julien Mairal, Priya Goyal, Piotr Bojanowski, and
  Armand Joulin.
\newblock Unsupervised learning of visual features by contrasting cluster
  assignments.
\newblock {\em arXiv preprint arXiv:2006.09882}, 2020.

\bibitem{caron2021emerging}
Mathilde Caron, Hugo Touvron, Ishan Misra, Herv{\'e} J{\'e}gou, Julien Mairal,
  Piotr Bojanowski, and Armand Joulin.
\newblock Emerging properties in self-supervised vision transformers.
\newblock {\em ICCV}, 2021.

\bibitem{changpinyo2021conceptual}
Soravit Changpinyo, Piyush Sharma, Nan Ding, and Radu Soricut.
\newblock Conceptual 12m: Pushing web-scale image-text pre-training to
  recognize long-tail visual concepts.
\newblock In {\em Proceedings of the IEEE/CVF Conference on Computer Vision and
  Pattern Recognition}, pages 3558--3568, 2021.

\bibitem{chen2020simple}
Ting Chen, Simon Kornblith, Mohammad Norouzi, and Geoffrey Hinton.
\newblock A simple framework for contrastive learning of visual
  representations.
\newblock In {\em International conference on machine learning}, pages
  1597--1607. PMLR, 2020.

\bibitem{chen2021exploring}
Xinlei Chen and Kaiming He.
\newblock Exploring simple siamese representation learning.
\newblock In {\em CVPR}, 2021.

\bibitem{chen2021empirical}
Xinlei Chen, Saining Xie, and Kaiming He.
\newblock An empirical study of training self-supervised visual transformers.
\newblock {\em ICCV}, 2021.

\bibitem{chollet2016information}
Fran{\c{c}}ois Chollet.
\newblock Information-theoretical label embeddings for large-scale image
  classification.
\newblock {\em arXiv preprint arXiv:1607.05691}, 2016.

\bibitem{deng2009imagenet}
Jia Deng, Wei Dong, Richard Socher, Li-Jia Li, Kai Li, and Li Fei-Fei.
\newblock Imagenet: A large-scale hierarchical image database.
\newblock In {\em 2009 IEEE conference on computer vision and pattern
  recognition}, pages 248--255. Ieee, 2009.

\bibitem{deng2017marginal}
Jiankang Deng, Yuxiang Zhou, and Stefanos Zafeiriou.
\newblock Marginal loss for deep face recognition.
\newblock In {\em Proceedings of the IEEE Conference on Computer Vision and
  Pattern Recognition Workshops}, pages 60--68, 2017.

\bibitem{desai2021virtex}
Karan Desai and Justin Johnson.
\newblock Virtex: Learning visual representations from textual annotations.
\newblock In {\em Proceedings of the IEEE/CVF Conference on Computer Vision and
  Pattern Recognition}, pages 11162--11173, 2021.

\bibitem{devlin2018bert}
Jacob Devlin, Ming-Wei Chang, Kenton Lee, and Kristina Toutanova.
\newblock Bert: Pre-training of deep bidirectional transformers for language
  understanding.
\newblock {\em arXiv preprint arXiv:1810.04805}, 2018.

\bibitem{donahue2014decaf}
Jeff Donahue, Yangqing Jia, Oriol Vinyals, Judy Hoffman, Ning Zhang, Eric
  Tzeng, and Trevor Darrell.
\newblock Decaf: A deep convolutional activation feature for generic visual
  recognition.
\newblock In {\em International conference on machine learning}, pages
  647--655. PMLR, 2014.

\bibitem{dosovitskiy2020image}
Alexey Dosovitskiy, Lucas Beyer, Alexander Kolesnikov, Dirk Weissenborn,
  Xiaohua Zhai, Thomas Unterthiner, Mostafa Dehghani, Matthias Minderer, Georg
  Heigold, Sylvain Gelly, et~al.
\newblock An image is worth 16x16 words: Transformers for image recognition at
  scale.
\newblock {\em arXiv preprint arXiv:2010.11929}, 2020.

\bibitem{frome2013devise}
Andrea Frome, Greg~S Corrado, Jon Shlens, Samy Bengio, Jeff Dean, Marc'Aurelio
  Ranzato, and Tomas Mikolov.
\newblock Devise: A deep visual-semantic embedding model.
\newblock {\em Advances in neural information processing systems}, 26, 2013.

\bibitem{gordo2017beyond}
Albert Gordo and Diane Larlus.
\newblock Beyond instance-level image retrieval: Leveraging captions to learn a
  global visual representation for semantic retrieval.
\newblock In {\em Proceedings of the IEEE conference on computer vision and
  pattern recognition}, pages 6589--6598, 2017.

\bibitem{goyal2021self}
Priya Goyal, Mathilde Caron, Benjamin Lefaudeux, Min Xu, Pengchao Wang, Vivek
  Pai, Mannat Singh, Vitaliy Liptchinsky, Ishan Misra, Armand Joulin, et~al.
\newblock Self-supervised pretraining of visual features in the wild.
\newblock {\em arXiv preprint arXiv:2103.01988}, 2021.

\bibitem{goyal2019scaling}
Priya Goyal, Dhruv Mahajan, Abhinav Gupta, and Ishan Misra.
\newblock Scaling and benchmarking self-supervised visual representation
  learning.
\newblock In {\em ICCV}, 2019.

\bibitem{grill2020bootstrap}
Jean-Bastien Grill, Florian Strub, Florent Altch{\'e}, Corentin Tallec,
  Pierre~H Richemond, Elena Buchatskaya, Carl Doersch, Bernardo~Avila Pires,
  Zhaohan~Daniel Guo, Mohammad~Gheshlaghi Azar, et~al.
\newblock Bootstrap your own latent: A new approach to self-supervised
  learning.
\newblock {\em NeurIPS}, 2020.

\bibitem{he2020momentum}
Kaiming He, Haoqi Fan, Yuxin Wu, Saining Xie, and Ross Girshick.
\newblock Momentum contrast for unsupervised visual representation learning.
\newblock In {\em CVPR}, 2020.

\bibitem{he2017mask}
Kaiming He, Georgia Gkioxari, Piotr Doll{\'a}r, and Ross Girshick.
\newblock Mask r-cnn.
\newblock In {\em Proceedings of the IEEE international conference on computer
  vision}, pages 2961--2969, 2017.

\bibitem{he2016deep}
Kaiming He, Xiangyu Zhang, Shaoqing Ren, and Jian Sun.
\newblock Deep residual learning for image recognition.
\newblock In {\em Proceedings of the IEEE conference on computer vision and
  pattern recognition}, pages 770--778, 2016.

\bibitem{henaff2019data}
Olivier~J H{\'e}naff, Aravind Srinivas, Jeffrey De~Fauw, Ali Razavi, Carl
  Doersch, SM Eslami, and Aaron van~den Oord.
\newblock Data-efficient image recognition with contrastive predictive coding.
\newblock {\em arXiv preprint arXiv:1905.09272}, 2019.

\bibitem{spacy2}
Matthew Honnibal and Ines Montani.
\newblock {spaCy 2}: Natural language understanding with {B}loom embeddings,
  convolutional neural networks and incremental parsing.
\newblock To appear, 2017.

\bibitem{hu2018squeeze}
Jie Hu, Li Shen, and Gang Sun.
\newblock Squeeze-and-excitation networks.
\newblock In {\em Proceedings of the IEEE conference on computer vision and
  pattern recognition}, pages 7132--7141, 2018.

\bibitem{hudson2019gqa}
Drew~A Hudson and Christopher~D Manning.
\newblock Gqa: A new dataset for real-world visual reasoning and compositional
  question answering.
\newblock In {\em Proceedings of the IEEE/CVF conference on computer vision and
  pattern recognition}, pages 6700--6709, 2019.

\bibitem{jayaraman2014zero}
Dinesh Jayaraman and Kristen Grauman.
\newblock Zero shot recognition with unreliable attributes.
\newblock {\em arXiv preprint arXiv:1409.4327}, 2014.

\bibitem{jia2021scaling}
Chao Jia, Yinfei Yang, Ye Xia, Yi-Ting Chen, Zarana Parekh, Hieu Pham, Quoc~V
  Le, Yunhsuan Sung, Zhen Li, and Tom Duerig.
\newblock Scaling up visual and vision-language representation learning with
  noisy text supervision.
\newblock {\em arXiv preprint arXiv:2102.05918}, 2021.

\bibitem{khosla2020supervised}
Prannay Khosla, Piotr Teterwak, Chen Wang, Aaron Sarna, Yonglong Tian, Phillip
  Isola, Aaron Maschinot, Ce Liu, and Dilip Krishnan.
\newblock Supervised contrastive learning.
\newblock {\em arXiv preprint arXiv:2004.11362}, 2020.

\bibitem{kim2021vilt}
Wonjae Kim, Bokyung Son, and Ildoo Kim.
\newblock Vilt: Vision-and-language transformer without convolution or region
  supervision.
\newblock {\em arXiv preprint arXiv:2102.03334}, 2021.

\bibitem{kingma2014adam}
Diederik~P Kingma and Jimmy Ba.
\newblock Adam: A method for stochastic optimization.
\newblock {\em arXiv preprint arXiv:1412.6980}, 2014.

\bibitem{kolesnikov2020big}
Alexander Kolesnikov, Lucas Beyer, Xiaohua Zhai, Joan Puigcerver, Jessica Yung,
  Sylvain Gelly, and Neil Houlsby.
\newblock Big transfer (bit): General visual representation learning.
\newblock In {\em Computer Vision--ECCV 2020: 16th European Conference,
  Glasgow, UK, August 23--28, 2020, Proceedings, Part V 16}, pages 491--507.
  Springer, 2020.

\bibitem{krizhevsky2009learning}
Alex Krizhevsky, Geoffrey Hinton, et~al.
\newblock Learning multiple layers of features from tiny images.
\newblock 2009.

\bibitem{krizhevsky2012imagenet}
Alex Krizhevsky, Ilya Sutskever, and Geoffrey~E Hinton.
\newblock Imagenet classification with deep convolutional neural networks.
\newblock {\em Advances in neural information processing systems},
  25:1097--1105, 2012.

\bibitem{lecun1995convolutional}
Yann LeCun, Yoshua Bengio, et~al.
\newblock Convolutional networks for images, speech, and time series.

\bibitem{lecun1989handwritten}
Yann LeCun, Bernhard Boser, John Denker, Donnie Henderson, Richard Howard,
  Wayne Hubbard, and Lawrence Jackel.
\newblock Handwritten digit recognition with a back-propagation network.
\newblock {\em Advances in neural information processing systems}, 2, 1989.

\bibitem{li2021efficient}
Chunyuan Li, Jianwei Yang, Pengchuan Zhang, Mei Gao, Bin Xiao, Xiyang Dai, Lu
  Yuan, and Jianfeng Gao.
\newblock Efficient self-supervised vision transformers for representation
  learning.
\newblock {\em arXiv preprint arXiv:2106.09785}, 2021.

\bibitem{li2021align}
Junnan Li, Ramprasaath~R Selvaraju, Akhilesh~Deepak Gotmare, Shafiq Joty,
  Caiming Xiong, and Steven Hoi.
\newblock Align before fuse: Vision and language representation learning with
  momentum distillation.
\newblock {\em arXiv preprint arXiv:2107.07651}, 2021.

\bibitem{li2020oscar}
Xiujun Li, Xi Yin, Chunyuan Li, Pengchuan Zhang, Xiaowei Hu, Lei Zhang, Lijuan
  Wang, Houdong Hu, Li Dong, Furu Wei, et~al.
\newblock Oscar: Object-semantics aligned pre-training for vision-language
  tasks.
\newblock In {\em European Conference on Computer Vision}, pages 121--137.
  Springer, 2020.

\bibitem{lin2013network}
Min Lin, Qiang Chen, and Shuicheng Yan.
\newblock Network in network.
\newblock {\em arXiv preprint arXiv:1312.4400}, 2013.

\bibitem{lin2014microsoft}
Tsung-Yi Lin, Michael Maire, Serge Belongie, James Hays, Pietro Perona, Deva
  Ramanan, Piotr Doll{\'a}r, and C~Lawrence Zitnick.
\newblock Microsoft coco: Common objects in context.
\newblock In {\em European conference on computer vision}, pages 740--755.
  Springer, 2014.

\bibitem{liu2016large}
Weiyang Liu, Yandong Wen, Zhiding Yu, and Meng Yang.
\newblock Large-margin softmax loss for convolutional neural networks.
\newblock In {\em ICML}, volume~2, page~7, 2016.

\bibitem{liu2021Swin}
Ze Liu, Yutong Lin, Yue Cao, Han Hu, Yixuan Wei, Zheng Zhang, Stephen Lin, and
  Baining Guo.
\newblock Swin transformer: Hierarchical vision transformer using shifted
  windows.
\newblock {\em ICCV}, 2021.

\bibitem{lu2019vilbert}
Jiasen Lu, Dhruv Batra, Devi Parikh, and Stefan Lee.
\newblock Vilbert: Pretraining task-agnostic visiolinguistic representations
  for vision-and-language tasks.
\newblock {\em arXiv preprint arXiv:1908.02265}, 2019.

\bibitem{mensink2014costa}
Thomas Mensink, Efstratios Gavves, and Cees~GM Snoek.
\newblock Costa: Co-occurrence statistics for zero-shot classification.
\newblock In {\em Proceedings of the IEEE conference on computer vision and
  pattern recognition}, pages 2441--2448, 2014.

\bibitem{murphy2012machine}
Kevin~P Murphy.
\newblock {\em Machine learning: a probabilistic perspective}.
\newblock MIT press, 2012.

\bibitem{radford2021learning}
Alec Radford, Jong~Wook Kim, Chris Hallacy, Aditya Ramesh, Gabriel Goh,
  Sandhini Agarwal, Girish Sastry, Amanda Askell, Pamela Mishkin, Jack Clark,
  et~al.
\newblock Learning transferable visual models from natural language
  supervision.
\newblock {\em arXiv preprint arXiv:2103.00020}, 2021.

\bibitem{rumelhart1986learning}
David~E Rumelhart, Geoffrey~E Hinton, and Ronald~J Williams.
\newblock Learning representations by back-propagating errors.
\newblock {\em nature}, 323(6088):533--536, 1986.

\bibitem{sariyildiz2020learning}
Mert~Bulent Sariyildiz, Julien Perez, and Diane Larlus.
\newblock Learning visual representations with caption annotations.
\newblock In {\em European Conference on Computer Vision}, pages 153--170.
  Springer, 2020.

\bibitem{sharma2018conceptual}
Piyush Sharma, Nan Ding, Sebastian Goodman, and Radu Soricut.
\newblock Conceptual captions: A cleaned, hypernymed, image alt-text dataset
  for automatic image captioning.
\newblock In {\em Proceedings of the 56th Annual Meeting of the Association for
  Computational Linguistics (Volume 1: Long Papers)}, pages 2556--2565, 2018.

\bibitem{sohn2016improved}
Kihyuk Sohn.
\newblock Improved deep metric learning with multi-class n-pair loss objective.
\newblock In {\em Advances in neural information processing systems}, pages
  1857--1865, 2016.

\bibitem{su2019vl}
Weijie Su, Xizhou Zhu, Yue Cao, Bin Li, Lewei Lu, Furu Wei, and Jifeng Dai.
\newblock Vl-bert: Pre-training of generic visual-linguistic representations.
\newblock {\em arXiv preprint arXiv:1908.08530}, 2019.

\bibitem{szegedy2015going}
Christian Szegedy, Wei Liu, Yangqing Jia, Pierre Sermanet, Scott Reed, Dragomir
  Anguelov, Dumitru Erhan, Vincent Vanhoucke, and Andrew Rabinovich.
\newblock Going deeper with convolutions.
\newblock In {\em Proceedings of the IEEE conference on computer vision and
  pattern recognition}, pages 1--9, 2015.

\bibitem{szegedy2016rethinking}
Christian Szegedy, Vincent Vanhoucke, Sergey Ioffe, Jon Shlens, and Zbigniew
  Wojna.
\newblock Rethinking the inception architecture for computer vision.
\newblock In {\em Proceedings of the IEEE conference on computer vision and
  pattern recognition}, pages 2818--2826, 2016.

\bibitem{thomee2016yfcc100m}
Bart Thomee, David~A Shamma, Gerald Friedland, Benjamin Elizalde, Karl Ni,
  Douglas Poland, Damian Borth, and Li-Jia Li.
\newblock Yfcc100m: The new data in multimedia research.
\newblock {\em Communications of the ACM}, 59(2):64--73, 2016.

\bibitem{tian2019contrastive}
Yonglong Tian, Dilip Krishnan, and Phillip Isola.
\newblock Contrastive multiview coding.
\newblock {\em arXiv preprint arXiv:1906.05849}, 2019.

\bibitem{tian2020makes}
Yonglong Tian, Chen Sun, Ben Poole, Dilip Krishnan, Cordelia Schmid, and
  Phillip Isola.
\newblock What makes for good views for contrastive learning.
\newblock {\em arXiv preprint arXiv:2005.10243}, 2020.

\bibitem{touvron2021training}
Hugo Touvron, Matthieu Cord, Matthijs Douze, Francisco Massa, Alexandre
  Sablayrolles, and Herv{\'e} J{\'e}gou.
\newblock Training data-efficient image transformers \& distillation through
  attention.
\newblock In {\em International Conference on Machine Learning}, pages
  10347--10357. PMLR, 2021.

\bibitem{van2008visualizing}
Laurens Van~der Maaten and Geoffrey Hinton.
\newblock Visualizing data using t-sne.
\newblock {\em Journal of machine learning research}, 2008.

\bibitem{vaswani2017attention}
Ashish Vaswani, Noam Shazeer, Niki Parmar, Jakob Uszkoreit, Llion Jones,
  Aidan~N Gomez, {\L}ukasz Kaiser, and Illia Polosukhin.
\newblock Attention is all you need.
\newblock In {\em Advances in neural information processing systems}, pages
  5998--6008, 2017.

\bibitem{wang2018learning}
Liwei Wang, Yin Li, Jing Huang, and Svetlana Lazebnik.
\newblock Learning two-branch neural networks for image-text matching tasks.
\newblock {\em IEEE Transactions on Pattern Analysis and Machine Intelligence},
  41(2):394--407, 2018.

\bibitem{wang2016learning}
Liwei Wang, Yin Li, and Svetlana Lazebnik.
\newblock Learning deep structure-preserving image-text embeddings.
\newblock In {\em Proceedings of the IEEE conference on computer vision and
  pattern recognition}, pages 5005--5013, 2016.

\bibitem{wang2021pyramid}
Wenhai Wang, Enze Xie, Xiang Li, Deng-Ping Fan, Kaitao Song, Ding Liang, Tong
  Lu, Ping Luo, and Ling Shao.
\newblock Pyramid vision transformer: A versatile backbone for dense prediction
  without convolutions.
\newblock {\em arXiv preprint arXiv:2102.12122}, 2021.

\bibitem{wang2018zero}
Xiaolong Wang, Yufei Ye, and Abhinav Gupta.
\newblock Zero-shot recognition via semantic embeddings and knowledge graphs.
\newblock In {\em Proceedings of the IEEE conference on computer vision and
  pattern recognition}, pages 6857--6866, 2018.

\bibitem{wang2021simvlm}
Zirui Wang, Jiahui Yu, Adams~Wei Yu, Zihang Dai, Yulia Tsvetkov, and Yuan Cao.
\newblock Simvlm: Simple visual language model pretraining with weak
  supervision.
\newblock {\em arXiv preprint arXiv:2108.10904}, 2021.

\bibitem{wu2021cvt}
Haiping Wu, Bin Xiao, Noel Codella, Mengchen Liu, Xiyang Dai, Lu Yuan, and Lei
  Zhang.
\newblock Cvt: Introducing convolutions to vision transformers.
\newblock {\em arXiv preprint arXiv:2103.15808}, 2021.

\bibitem{wu2019detectron2}
Yuxin Wu, Alexander Kirillov, Francisco Massa, Wan-Yen Lo, and Ross Girshick.
\newblock Detectron2.
\newblock \url{https://github.com/facebookresearch/detectron2}, 2019.

\bibitem{xian2016latent}
Yongqin Xian, Zeynep Akata, Gaurav Sharma, Quynh Nguyen, Matthias Hein, and
  Bernt Schiele.
\newblock Latent embeddings for zero-shot classification.
\newblock In {\em Proceedings of the IEEE conference on computer vision and
  pattern recognition}, pages 69--77, 2016.

\bibitem{xian2017zero}
Yongqin Xian, Bernt Schiele, and Zeynep Akata.
\newblock Zero-shot learning-the good, the bad and the ugly.
\newblock In {\em Proceedings of the IEEE Conference on Computer Vision and
  Pattern Recognition}, pages 4582--4591, 2017.

\bibitem{yang2021focal}
Jianwei Yang, Chunyuan Li, Pengchuan Zhang, Xiyang Dai, Bin Xiao, Lu Yuan, and
  Jianfeng Gao.
\newblock Focal attention for long-range interactions in vision transformers.
\newblock {\em Advances in Neural Information Processing Systems}, 34, 2021.

\bibitem{yuan2021florence}
Lu Yuan, Dongdong Chen, Yi-Ling Chen, Noel Codella, Xiyang Dai, Jianfeng Gao,
  Houdong Hu, Xuedong Huang, Boxin Li, Chunyuan Li, et~al.
\newblock Florence: A new foundation model for computer vision.
\newblock {\em arXiv preprint arXiv:2111.11432}, 2021.

\bibitem{yun2019cutmix}
Sangdoo Yun, Dongyoon Han, Seong~Joon Oh, Sanghyuk Chun, Junsuk Choe, and
  Youngjoon Yoo.
\newblock Cutmix: Regularization strategy to train strong classifiers with
  localizable features.
\newblock In {\em Proceedings of the IEEE/CVF International Conference on
  Computer Vision}, pages 6023--6032, 2019.

\bibitem{zellers2019recognition}
Rowan Zellers, Yonatan Bisk, Ali Farhadi, and Yejin Choi.
\newblock From recognition to cognition: Visual commonsense reasoning.
\newblock In {\em Proceedings of the IEEE/CVF Conference on Computer Vision and
  Pattern Recognition}, pages 6720--6731, 2019.

\bibitem{zhang2017mixup}
Hongyi Zhang, Moustapha Cisse, Yann~N Dauphin, and David Lopez-Paz.
\newblock mixup: Beyond empirical risk minimization.
\newblock {\em arXiv preprint arXiv:1710.09412}, 2017.

\bibitem{zhang2021multi}
Pengchuan Zhang, Xiyang Dai, Jianwei Yang, Bin Xiao, Lu Yuan, Lei Zhang, and
  Jianfeng Gao.
\newblock Multi-scale vision longformer: A new vision transformer for
  high-resolution image encoding.
\newblock In {\em Proceedings of the IEEE/CVF International Conference on
  Computer Vision}, pages 2998--3008, 2021.

\bibitem{zhang2021vinvl}
Pengchuan Zhang, Xiujun Li, Xiaowei Hu, Jianwei Yang, Lei Zhang, Lijuan Wang,
  Yejin Choi, and Jianfeng Gao.
\newblock Vinvl: Revisiting visual representations in vision-language models.
\newblock In {\em Proceedings of the IEEE/CVF Conference on Computer Vision and
  Pattern Recognition}, pages 5579--5588, 2021.

\bibitem{zhang2020contrastive}
Yuhao Zhang, Hang Jiang, Yasuhide Miura, Christopher~D Manning, and Curtis~P
  Langlotz.
\newblock Contrastive learning of medical visual representations from paired
  images and text.
\newblock {\em arXiv preprint arXiv:2010.00747}, 2020.

\end{thebibliography}
}

\newpage
\appendix

\begin{table*}[t!]
  \centering
    \setlength{\tabcolsep}{2.2pt}
  \scalebox{0.86}{
\begin{tabular}{c | c c c c c c | c c } 
 \toprule
 Dataset & \#Concepts & Vocab. Size & Train size & Test size & Evaluation metric & Source link & Linear Probe & Zero-shot \\ 
 \midrule
 Food-101 & 102 & 139 & 75,750 & 25,250 & Accuracy & \href{https://www.tensorflow.org/datasets/catalog/food101}{Tensorflow} & \checkmark & \checkmark \\ 
 CIFAR-10 & 10 & 10 & 50,000 & 10,000 & Accuracy & \href{https://www.tensorflow.org/datasets/catalog/cifar10}{TensorFlow}  &  \checkmark & \checkmark \\
 CIFAR-100 & 100 & 100 & 50,000 & 10,000 & Accuracy & \href{https://www.tensorflow.org/datasets/catalog/cifar100}{TensorFlow}  & \checkmark & \checkmark \\
SUN397 & 397 & 432 & 19,850 & 19,850 & Accuracy & \href{https://www.tensorflow.org/datasets/catalog/sun397}{Tensorflow}  & \checkmark & \\
Stanford Cars & 196 & 291 & 8,144 & 8,041 & Accuracy & \href{https://ai.stanford.edu/~jkrause/cars/car_dataset.html}{Stanfold Cars}   &  \checkmark & \\
FGVC Aircraft (variants) & 100 & 115 & 6,667 & 3,333 & Mean-per-class & \href{https://www.robots.ox.ac.uk/~vgg/data/fgvc-aircraft/}{FGVC website}  & \checkmark & \\
VOC2007 classification & 20 & 20 & 5,011 & 4,952& 11-point mAP & \href{http://host.robots.ox.ac.uk/pascal/VOC/voc2007/index.html}{voc2007}  & \checkmark & \checkmark \\
Describable Textures & 47 & 47 & 3,760& 1,880& Accuracy& \href{https://www.tensorflow.org/datasets/catalog/dtd}{TensorFlow}  & \checkmark & \checkmark \\
Oxford-IIIT  Pets & 37 & 53 & 3,680& 3,669& Mean-per-class& \href{https://www.robots.ox.ac.uk/~vgg/data/pets/}{Oxford-IIIT Pet}  & \checkmark & \checkmark \\
Caltech-101& 102 & 122 & 3,060& 6084& Mean-per-class& \href{https://www.tensorflow.org/datasets/catalog/caltech101}{TensorFlow}  & \checkmark & \checkmark \\
Oxford Flowers 102& 102& 147 & 2,040& 6,149 & Mean-per-class& \href{https://www.tensorflow.org/datasets/catalog/oxford_flowers102}{TensorFlow}  & \checkmark & \checkmark \\
MNIST& 10 & 10 & 60,000& 10,000 & Accuracy& \href{https://www.tensorflow.org/datasets/catalog/mnist}{TensorFlow}  & \checkmark & \\
 FER~2013~$^{\ast}$& 8& 12 & 32,298& 3,589 & Accuracy& \href{https://www.kaggle.com/c/challenges-in-representation-learning-facial-expression-recognition-challenge/data}{Kaggle fer2013}  & \checkmark & \checkmark \\
STL10 & 10 & 10 & 5,000& 8,000& Accuracy& \href{https://www.tensorflow.org/datasets/catalog/stl10}{TensorFlow}  & \checkmark & \\
GTSRB $^{\ast}$& 43 & 85 & 26,728 & 12,630 & Accuracy & \href{https://benchmark.ini.rub.de/gtsrb_dataset.html}{GTSRB website}  & \checkmark & \\
PatchCamelyon & 2 & 6 & 294,912 & 32,768 & Accuracy & \href{https://www.tensorflow.org/datasets/catalog/patch_camelyon}{TensorFlow}  & \checkmark & \checkmark \\
UCF101 $^{\ast}$ & 101 & 153 & 9,537 & 3783 & Accuracy & \href{https://www.tensorflow.org/datasets/catalog/ucf101}{TensorFlow}  & \checkmark & \\
Hateful Memes & 2 & 2 & 8,500 & 500 & ROC-AUC & \href{https://ai.facebook.com/blog/hateful-memes-challenge-and-data-set/}{FaceBook}  & \checkmark & \checkmark \\
EuroSAT &	10	 & 20 & 5,000  & 5,000 &	Accuracy &
\href{https://www.tensorflow.org/datasets/catalog/eurosat}{TensorFlow}  &  & \checkmark \\
Resisc45 &	45	& 59 & 3,150	& 25,200 &	Accuracy &
\href{https://www.tensorflow.org/datasets/catalog/resisc45}{TensorFlow}  &   & \checkmark \\
Rendered-SST2&	2&	2 & 6,920 &	1,821 &	Accuracy &  \href{https://github.com/openai/CLIP/blob/main/data/rendered-sst2.md}{OpenAI}  &   & \checkmark \\

\bottomrule
\end{tabular}
}
\vspace{-2mm}
\caption{Statistics of datasets used in zero-shot and linear probe.  $^{\ast}$ indicates dataset whose train/test size we obtained is slightly different from Table 9 in~\cite{radford2021learning}. \checkmark indicates the dataset is used in this setting.  The datasets are chosen based on the criterion if we can reproduce the numbers reported from~\cite{radford2021learning} and their availability. }
\label{table:downstream_ic_dataset}
\end{table*}

\section{Validation dataset details}
\label{sec:validation_datasets}

In addition to the training datasets listed in our main submission, we list in Table~\ref{table:downstream_ic_dataset} the statistics for all the validation datasets used in our experiments. Similar to the Table 1 in our main submission, we calculate the vocabulary size for each dataset, which is typically more than the number of concepts (classes).


\section{Experiment details}

\subsection{Training on image classification data}

This part mainly explains the detailed experiment setups for Sec.~4.1 in our main submission. 

\xhdr{Model architecture}. We employ two representative architectures,  ResNet~\cite{he2016deep} and Swin Transformer~\cite{liu2021Swin} to build the visual encoder.  The globally pooled feature from last visual encoder layer is used as the visual feature.
For language encoder, we use a 12-layer Transformer~\cite{vaswani2017attention} with hidden dimension of 512 following~\cite{radford2021learning}. Features from visual and textual encoder are projected to the same dimension of 512, using two linear projection layers.

\noindent \textbf{Training protocol}. For optimization, we use SGD~\cite{rumelhart1986learning} for all CNN models, while AdamW~\cite{kingma2014adam} for all models with Transformers on either vision or language side. We set the learning rate to 0.4 and 0.002, weight decay to 1e-4 and 0.05 for SGD and AdamW optimizer, respectively. All models are trained for 500 epochs with a batch size of 4096. We use same set of data augmentation and regularization as in~\cite{liu2021Swin}, but do not use MixUp~\cite{zhang2017mixup} and CutMix~\cite{yun2019cutmix} except for the last column in Table 2 of our main submission. For all training, we used a cosine learning rate schedule, with 5 epochs and 20 epochs warmup for ResNet and Swin Transformer, respectively.

\subsection{Training on image-text-label space}

\xhdr{Training protocol for Sec.~4.2.1}. We use Swin-Tiny as the visual encoder and follow the training settings in Section 4.1  mostly to train the models on the joint of image-label and image-text pairs. However, we notice there is a severe imbalance between image-label and image-text data as shown in Table~1 in our main submission (\textit{e.g.}, there are around 1.3M images in ImageNet-1K while above 12M images in GCC-12M dataset). To ensure that the model training is not biased to the dominant image-text pairs, we develop a balanced data sampler for two data types. More specifically, at each epoch, we randomly sample a subset of image-text pairs that has the equal or similar size to that of image-label data. In this case, the model sees half image-label data and half image-text data at each iteration for a balanced learning. We keep the number of training epochs the same as 500, so the effective number of training epochs on the image-text dataset is roughly 500$\times$(size of image-label dataset)/(size of image-text pair dataset). For example, the model learns from GCC-12M for around 40 epochs. We find this balanced sampling strategy is very important to achieve the reported performance in our main submission.

\xhdr{Training protocol for Sec.~4.2.2}. We followed the training protocol in CLIP~\cite{radford2021learning} for fair comparison. Specifically, we merely used random crop for dataset augmentation in all model trainings. All models including the baseline models are trained for 32 epochs, with batch size 4096, initial learning rate 1e-3 and weight decay 0.1. We also used a cosine learning rate scheduler with 5000 warmup iterations. 

\section{More results}

\subsection{Results over separate datasets}

\begin{figure*}[t!]
    \centering
    \noindent\includegraphics[width=0.96\textwidth]{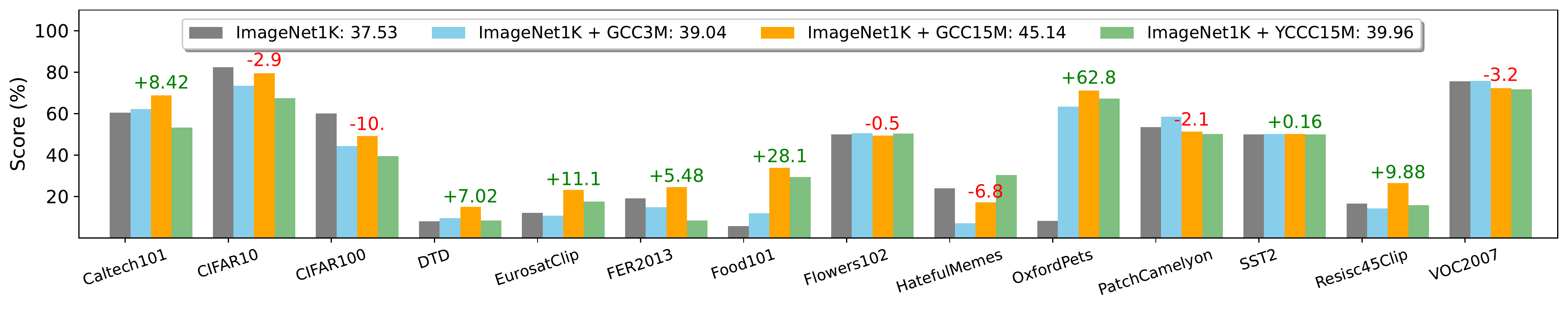}    
    \vspace{-4mm}
    \caption{Zero-shot classification on 14 datasets by adding different image-caption pairs into the ImageNet-1K. The averaged scores of each method is reported in the legend. The gain of \oursloss{} on mixed data (ImageNet1K+GCC15M) over image-label data (ImageNet1K) is shown.}
    \label{fig:transfer_zs_cls_1k}
    \vspace{-3mm}
\end{figure*}

In Figure~\ref{fig:transfer_zs_cls_1k}, we show the zero-shot classification on 14 datasets by adding different image-caption pairs into the ImageNet-1K, \ie the methods compared in Table 5 in the main text.  \oursloss{} takes the advantages of learning rich concept coverage from image-text pairs: On most of the datasets, it outperforms the baseline, especially on fine-grained classification tasks such as Food101 and OxfordPets.

\subsection{Results with larger vision backbone}

In our main submission, we used Swin-Tiny as the visual backbone to study how our \oursloss{} perform when trained on the combination of image-label data ImageNet-21K and image-text pairs YFCC-14M in Table 6. Here, we investigate whether increase the capacity of the vision backbone can further improve the representation learning.

As shown in Table~\ref{tab:perf_with_swin_base}, we observe consistent trend as in Table 6 of our main submission. Though using similar amount of image-text-label corpus, combining two type of data can \emph{significantly} improve the zero-shot recognition performance on both ImageNet-1K (\textbf{8.6} points) and other 14 datasets in average (\textbf{11.0} points). When using the full set of ImageNet-21K and YFCC-14M, both performance can be further improved significantly. These results suggest that our method is agnostic to different model sizes and thus a generic learning paradigm for visual-semantic representations. For comparison, we also list the numbers for Swin-Tiny models after each ``/''. Clearly, increasing the visual encoder size brings substantial gains in all cases, and particularly significant for the combination of both data types.

\begin{table}[t]
    \centering
    \footnotesize
    \setlength{\tabcolsep}{2.4pt}
  \scalebox{0.95}{
    \begin{tabular}{lcccc}
    \toprule
        \multirow{2}{*}{Training Data} & \multirow{2}{*}{Method} & \multicolumn{2}{c}{Zero-shot}  \\
        \cmidrule{3-4}
        & & ImageNet-1K & 14 datasets   \\
        \midrule
        YFCC-14M & CLIP & 32.4/30.1 & 37.5/36.3 \\
        ImageNet-21K & \oursloss & 29.9/28.5 &  42.4/37.8 \\
        YFCC-14M(half)+ImageNet-21K(half) & \oursloss & 41.0/36.4 & 48.5/45.5 \\           
        YFCC-14M+ImageNet-21K & \oursloss & \textbf{43.8}/\textbf{40.5} & \textbf{52.2}/\textbf{49.1}  \\     
        \bottomrule
    \end{tabular}}
    \vspace{-3mm}
    \caption{Ablation studies on the training datasets and tasks. We use Swin-Base~\cite{liu2021Swin} as the vision backbone. Each model is pre-trained with 32 epochs following CLIP~\cite{radford2021learning}. Numbers before and after each ``/'' are with Swin-Base and Swin-Tiny, respectively.}
    \label{tab:perf_with_swin_base}
      \vspace{-3mm}
\end{table}


\subsection{Transfer to object detection}

In the Table 5 of our main submission, we mainly studied whether image-text pairs can bring benefits to object detection transfer learning compared with the models solely trained on image-label data. As we demonstrated in Table 6 of our main submission, image-label data can help to learn more discriminative representations, and thus benefits ImageNet-1K finetuning and linear probing. Here, we further study whether the learned representations can generalize to object detection task as well. Specifically, we use the Swin-Tiny models pretrained in Table 6 as the vision backbones and train a Mask R-CNN model with 1$\times$ schedule following the default settings in Swin Transformer~\cite{liu2021Swin} based on Detectron2~\cite{wu2019detectron2}. 
In Table~\ref{tab:perf_object_detection}, we can see combining two data types with similar amount clearly improve the object detection performance by around 2 points for both box and mask mAP, compared with the CLIP-based model trained on YFCC-14M. This further validates our note that representations learned from pure image-text pair data usually lack the discirminative ability required by transfer learning to image recognition and object detection. As expected, using the full set (last row) brings further around 1 point improvement for both metrics. Along with the reported numbers in Table 5 of our main submission, these results together imply that adding image-text pairs to image-label data and the other way around can universally help to learn a better visual representations compared with the individual counterparts. Adding image-text pairs data can enrich and smoothen the semantic space which may implicitly prompt distinctive representations for the concepts in COCO object detection, while adding image-label data directly imposes the pressure to learn more discriminative representations.

\begin{table}[t]
    \centering
    \footnotesize
    \setlength{\tabcolsep}{2.6pt}
    \begin{tabular}{lcccc}
    \toprule
        \multirow{2}{*}{Training Data} & \multirow{2}{*}{Method} & \multicolumn{2}{c}{Object Detection}  \\
        \cmidrule{3-4}
        & & box mAP & mask mAP \\
        \midrule
        YFCC-14M & CLIP & 39.9 & 37.3 \\
        ImageNet-21K & \oursloss & 41.4 & 38.6 \\
        YFCC-14M(half)+ImageNet-21K(half) & \oursloss & 41.9 & 39.0 \\           
        YFCC-14M+ImageNet-21K & \oursloss & \textbf{43.1} & \textbf{40.0}  \\     
        \bottomrule
    \end{tabular}
    \vspace{-3mm}
    \caption{Object detection transfer learning with different models. We use the pretrained Swin-Tiny models listed in Table 6 of our main submission as the vision backbone.}
    \label{tab:perf_object_detection}
      \vspace{-3mm}
\end{table}


\section{More analysis}

\subsection{Concept distribution}

\begin{figure*}[t]
    \centering
    \noindent\includegraphics[height=0.25\textwidth]{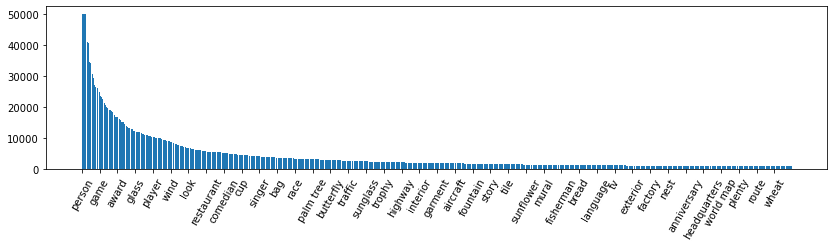}       
    \vspace{-2mm}
    \noindent\includegraphics[height=0.27\textwidth]{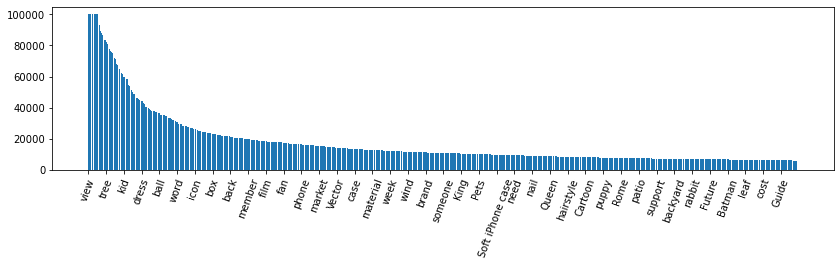}       
    \noindent\includegraphics[height=0.255\textwidth]{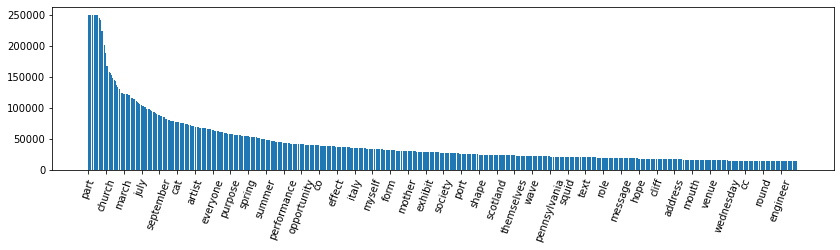}       
    \vspace{-2mm}
    \caption{From top to bottom, bar charts are top 1000 most frequent concepts in GCC-3M, GCC-12M and YFCC-14M, respectively. We trim the heights of most frequent concepts for better display. For clarity, we display the concept name for every 25 concepts.}
    \label{fig:concept_dists}
    \vspace{-3mm}
\end{figure*}

The concepts residing in the training data is arguably crucial to the model learning. Both CLIP~\cite{radford2021learning} and ALIGN~\cite{jia2021scaling} exhaustively collect hundreds of millions of image-text pairs to cover as many visual concepts as possible. Though the datasets used in our experiments are at much smaller scale, we are still interested in the concept distributions of different datasets. In Fig.~\ref{fig:concept_dists}, we show the occurrences of top 1000 concepts in GCC-3M, GCC-12M and YFCC-14M. Along with the remaining concepts that do not show here, all three datasets have extreme long-tail distributions. For example, the most frequent concept ``view'' in GCC-12M appears over 185,363 times, while the 10k-th concept ``candle holder'' only appears 501 times, knowing that there are more than 584k concepts in the whole set.

Interestingly, we find the overlap of most common concepts across three datasets is lower than what we expect. Table~\ref{tab:concet_overlap_ratio} shows the overlap ratios of top 10k concepts among three datasets. These relatively lower overlapping indicates the sufficient diversities and complementary among them. 

\begin{table}[t]
    \centering
    \footnotesize
    \setlength{\tabcolsep}{2.6pt}
    \begin{tabular}{l|cccc}
    \toprule
    Dataset & GCC-3M & GCC-12M & YFCC-14M \\
    \midrule
    GCC-3M  & 100\%  &  46.5\%       &   50.2\%       \\
    GCC-12M  & 46.5\%  &    100\%     &  37.9\%        \\
    YFCC-14M  & 50.2\% &     37.9\%    &   100\%       \\
    \bottomrule
    \end{tabular}
    \vspace{-3mm}
    \caption{Overlap ratio of top 10k concepts among GCC-3M, GCC-12M and YFCC-14M. The matrix is symmetric.}
    \label{tab:concet_overlap_ratio}
    \vspace{-3mm}
\end{table}

\subsection{Concept coverage}

Given the concept distributions above, we further investigate the concept coverage between training datasets and validation datasets. In Table~\ref{tab:dataset_coverage}, we calculate the coverage ratio to be the percentage of concepts mentioned by the pretraining data, including ImageNet-1K, ImageNet-21K, GCC-3M, GCC-12M and YFCC-14M. Coverage ratios equal or larger than \textbf{50}\% are highlighted.

Accordingly, for image-label dataset ImageNet-1K, it has some overlaps with CIFAR-100 (24.0\%) and Caltech-101 (24.5\%). This may explain why the zero-shot performance on these two datasets shown in Fig.~\ref{fig:transfer_zs_cls_1k} is relative higher. In contrast, we also notice that even with less or no coverage, the model pretrained on ImageNet-1K with our method still attain reasonably good zero-shot performance on datasets like CIFAR-10, Flowers102, Oxford Pet, \textit{etc}.

Similarly, for ImageNet-21K, it covers a certain proportion of concepts in the validation sets, such as CIFAR-10, CIFAR-100, Caltech-101, \textit{etc}, and we did observe high zero-shot recognition performance on them in the Table 6 of our main submission. Nevertheless, for other datasets like \textit{Hateful Memes}, \textit{PatchCamelyon}, there are zero concept overlaps, while our model still realizes reasonable performance. This indicates that our model is not just memorizing the concepts appearing in the training datasets, but also learns to understand the underlying structures of different concepts, which has been also demonstrated in Fig.~5 of our main submission.

Finally, we find image-text pairs data have higher coverage of concepts than image-label datasets almost on all validation sets. Among the three image-text pair datasets, GCC-12M has relatively higher coverage than the other two datasets. This may also explain why we observe better performance in the comparisons shown in Table 5 of our main submission. However, we also notice that higher concept coverage does not necessarily means better zero-shot performance. For example, even though all of these three datasets have a fully coverage of concepts in CIFAR-10 and CIFAR-100, adding them into the pretraining hurts the performance as shown in Fig.~\ref{fig:transfer_zs_cls_1k}. We suspect there might be some significant gaps in the image domain between the pretraining and validation datasets even though they share common semantic concepts. Moreover, images in image-text pairs usually contain multiple objects, the coverage of concepts does not necessarily means the model can learn to grounding the concepts to the specific image contents. How to better leverage the image-text pair data and build a more gounded visual understanding worth further studies.

\begin{table*}[t!]
  \centering
    \setlength{\tabcolsep}{2.4pt}
  \scalebox{0.86}{
\begin{tabular}{c c c | c c c c c c c c c c} 
 \toprule
 \multicolumn{3}{c}{Dataset} & \multicolumn{2}{c}{ImageNet-1K} & \multicolumn{2}{c}{ImageNet-21K} & \multicolumn{2}{c}{GCC-3M} & \multicolumn{2}{c}{GCC-12M} & \multicolumn{2}{c}{YFCC-14M} \\ 
 \midrule
 Name & \#Concepts & Vocab. Size & Cover. & \#Img/C. & Cover. & \#Img/C. & Cover. & \#Img/C. & Cover. & \#Img/C. & Cover. & \#Img/C. \\
 \midrule
 ImageNet-1K  & 1,000 & 1,233 & \textbf{100}\% & 1300 & 0\% & 0 & 45.3\% & 247.0 & \textbf{78.5}\% & 851.1 & \textbf{69.3}\% & 1930.8 \\
 Food-101 & 102 & 139 & 4.0\% & 1300.0 & 20.8\% & 650.0 & 21.8\% & 39.8 & \textbf{58.4}\% & 250.8 & \textbf{67.3}\% & 408.8 \\ 
 CIFAR-10 & 10 & 10 & 0.0\% & 0.0 & \textbf{90.0}\% & 650.0 & \textbf{100.0}\% & 6175.4 & \textbf{100.0}\% & 19969.8 & \textbf{100.0}\% & 32998.9 \\
 CIFAR-100 & 100 & 100 & 24.0\% & 1300.0 & \textbf{65.0}\% & 650.0 & \textbf{95.0}\% & 3928.4 & \textbf{99.0}\% & 15628.5 & \textbf{99.0}\% & 18303.2 \\
SUN397 & 397 & 432 & 5.0\% & 1300.0 & 28.5\%& 650.0 & 48.1\% & 818.9 & \textbf{65.5}\% & 2355.4 & \textbf{66.5}\% & 7043.2 \\
Stanford Cars & 196 & 291 & 0.0\% & 0.0 & 0.0\% & 0.0 & 0.0\% & 0.0 & 0.0\% & 0.0 & 0.0\% & 0.0 \\
FGVC Aircraft (variants) & 100 & 115 & 0.0\% & 0.0 & 0.0\% & 0.0 & 0.0\% & 0.0 & 22.0\% & 4.1 & 0.0\% & 0.0 \\
VOC2007 classification & 20 & 20 & 0.0\% & 0.0 & 75.0\% & 650.0 & \textbf{85.0}\% & 14721.6 & \textbf{85.0}\% & 19934.8 & \textbf{85.0}\% & 31448.8 \\
Describable Textures & 47 & 47 & 0.0\% & 0.0 & 4.3\% & 650.0 & 14.9\% & 8.9 & 27.7\% & 53.2 & 36.2\% & 181.7 \\
Oxford-IIIT  Pets & 37 & 53 & 5.4\% & 1300.0 & 13.5\% & 650.0 & 10.8\% & 80.9 & \textbf{64.9}\% & 134.0 & 37.8\% & 169.0 \\
Caltech-101& 102 & 122 & 24.5\% & 1300.0 & 43.1\% & 650.0 & \textbf{66.6}\% & 1633.8 & \textbf{83.3}\% & 5249.7 & \textbf{87.3}\% & 5017.7  \\
Oxford Flowers 102& 102& 147 & 10.0\% & 1300.0 & 40.2\% & 650.0 & 17.6\% & 53.2 & \textbf{50.0}\% & 194.3 & \textbf{65.7}\% & 422.7 \\
MNIST& 10 & 10 & 0.0\% & 0.0 & 0.0\% & 0.0 & 40.0\% & 0.8 & \textbf{100.0}\% & 46.0 & \textbf{90.0}\% & 68.8 \\
FER~2013~$^{\ast}$& 8& 12 & 0.0\% & 0.0 & 8.3\% & 650.0 & 25.0\% & 5.9 & 41.7\% & 29.2 & 41.7\% & 11.5 \\
STL10 & 10 & 10 & 0.0\%& 0.0 & 100\% & 650.0 & \textbf{100.0}\% & 8778.6 & \textbf{100.0}\% & 28547.6 & \textbf{100.0}\% & 45587.5 \\
GTSRB $^{\ast}$& 43 & 85 & 0.0\% & 0.0 & 0.0\% & 0.0 & 2.3\% & 12.7 & 2.3\% & 52.9 & 2.3\% & 551.3 \\
PatchCamelyon & 2 & 6 & 0.0\% & 0.0 & 0.0\% & 0.0 & 0.0\% & 0.0 & \textbf{50.0}\% & 143.0 & \textbf{50.0}\% & 15.0 \\
UCF101 $^{\ast}$ & 101 & 153 & 0.0\% & 0.0 & 0.0\% & 0.0 & 0.0\% & 0.0 & \textbf{51.5}\% & 66.4 & 0.0\% & 0.0 \\
Hateful Memes & 2 & 2 & 0.0\% & 0.0 & 0.0\% & 0.0 & \textbf{50.0}\% & 79.5 & \textbf{50.0}\% & 2742.5 & \textbf{50.0}\% & 321.5 \\
EuroSAT &	10	 & 20 & 0.0\% & 0.0 & 0.0\% & 0.0 &  20.0\% & 2946.6 & 30.0\% & 5266.3 & 30.0\% & 15458.7 \\
Resisc45 &	45	& 59 & 8.9\% & 1300.0 & 26.7\% & 650.0 & \textbf{71.1}\% & 3688.6 & \textbf{75.6}\% & 7572.0 & \textbf{80.0}\% & 26317.6 \\
Rendered-SST2&	2&	2 & 0.0\% & 0.0 & \textbf{50.0}\% & 650.0 & \textbf{50.0}\% & 1.0 & \textbf{100.0}\% & 114.0 & \textbf{100.0}\% & 1259.0 \\
\bottomrule
\end{tabular}
}
\vspace{-2mm}
\caption{Statistics of concept coverage between training and validation dataests. $^{\ast}$ indicates dataset whose train/test size we obtained is slightly different from Table 9 in~\cite{radford2021learning}. ``Cover.'' denotes the coverage ratio of concepts in target dataset by the training dataet. For those with non-zero coverage ratio, we also list the average number of images for each concept. For ImageNet-1K and ImageNet-21K, we estimate the number of images per concept by dividing the total number of images by total number of concepts, which are 1300 and 650, respectively}
\vspace{-4mm}
\label{tab:dataset_coverage}
\end{table*}

\subsection{Concept visualizations}
\begin{figure*}[t!]
    \centering
    \begin{minipage}{0.53\linewidth}
    \noindent\includegraphics[width=0.90\linewidth]{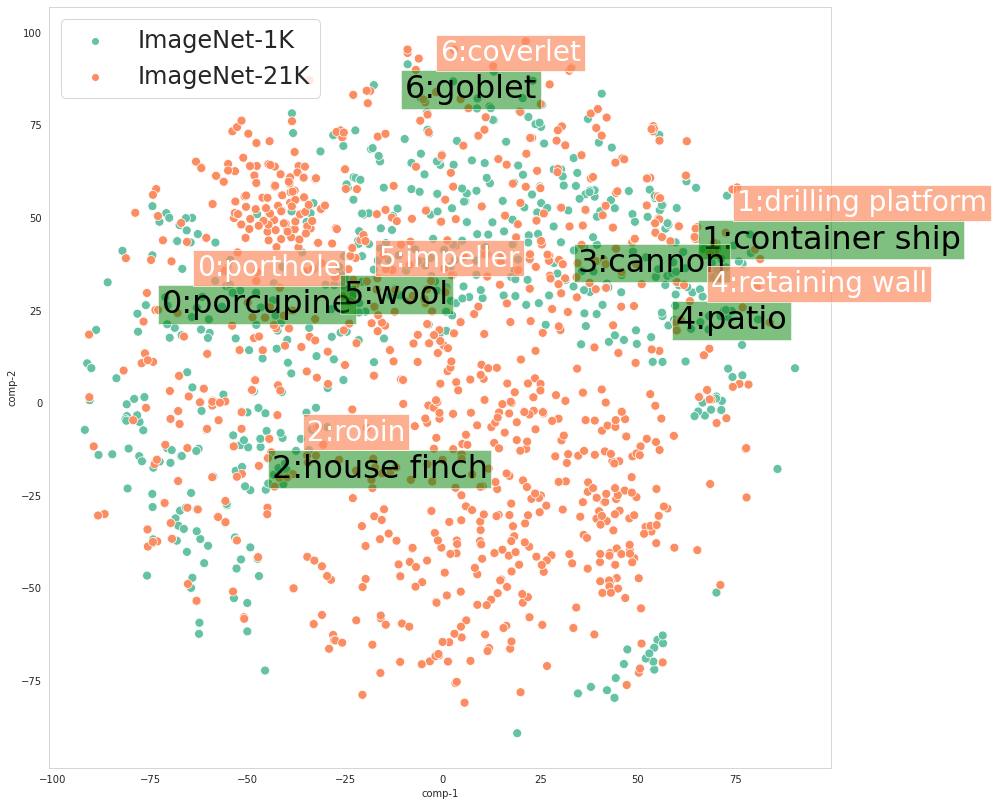}        
    \end{minipage}
    \begin{minipage}{0.46\linewidth}
    \noindent\includegraphics[width=0.99\linewidth]{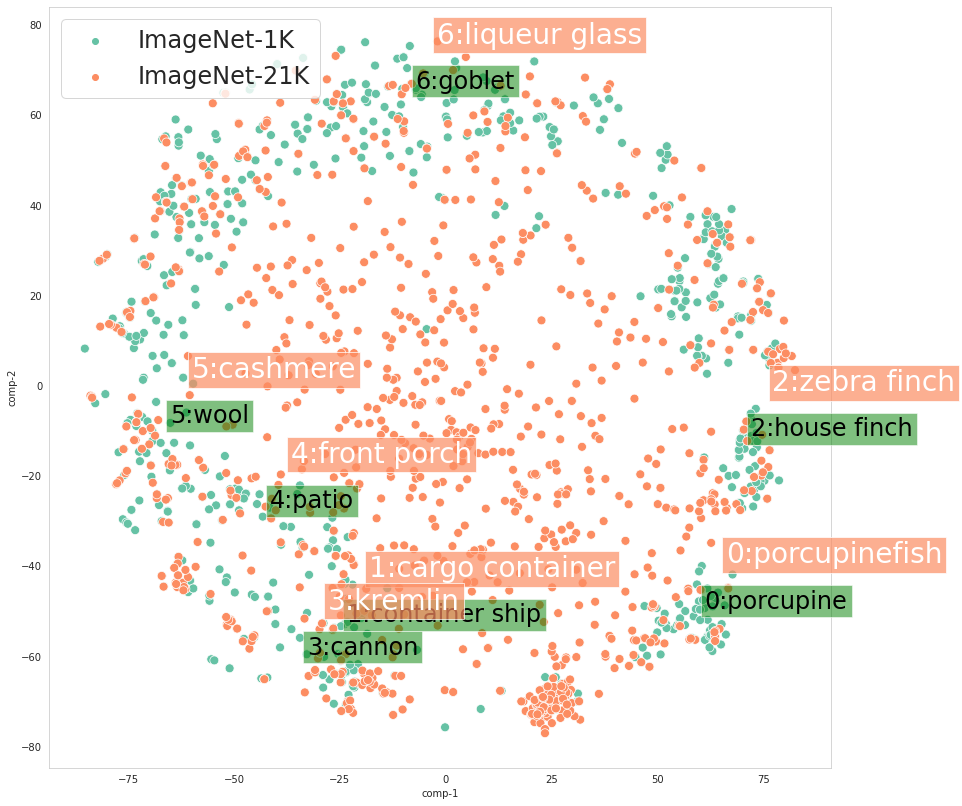}        
    \end{minipage}         
    \caption{Similar to Fig. 4 in our main submission, we further visualize the t-SNE embedding for visual concepts with models trained with ImageNet-1K (left) and ImageNet-1K+GCC-15M (right).}
    \label{fig:concept_tsne}
    \vspace{-3mm}
\end{figure*}

\begin{figure*}[t!]
    \centering
    \begin{minipage}{0.48\linewidth}
    \noindent\includegraphics[width=0.90\linewidth]{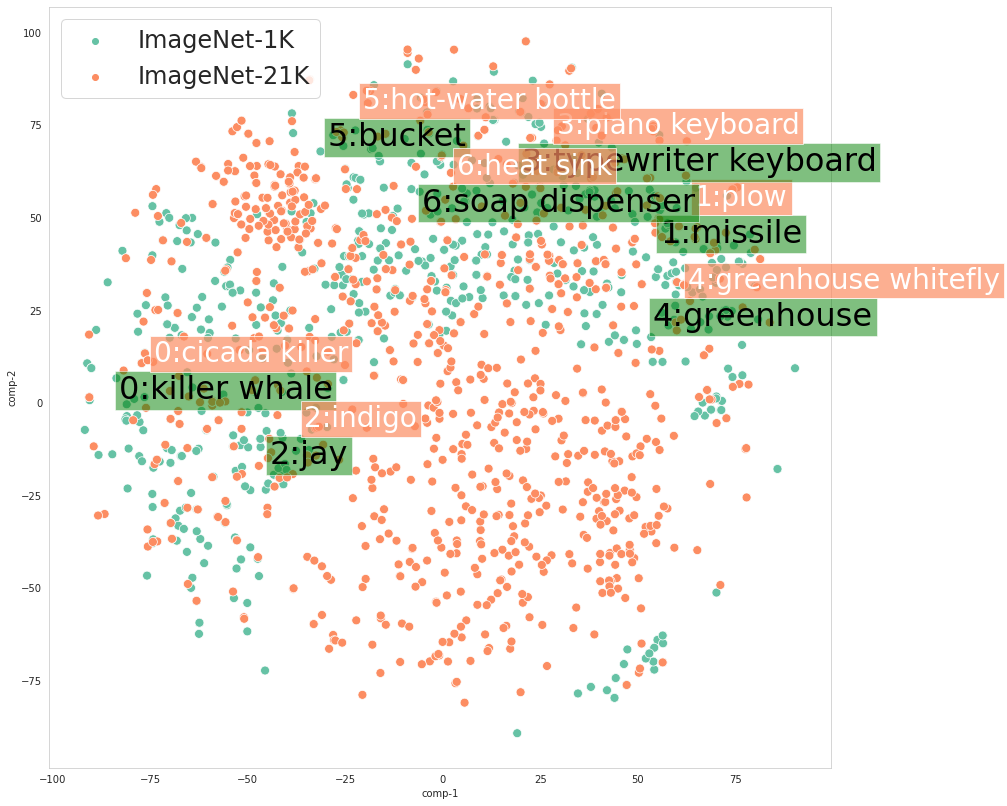}        
    \end{minipage}
    \begin{minipage}{0.50\linewidth}
    \noindent\includegraphics[width=0.99\linewidth]{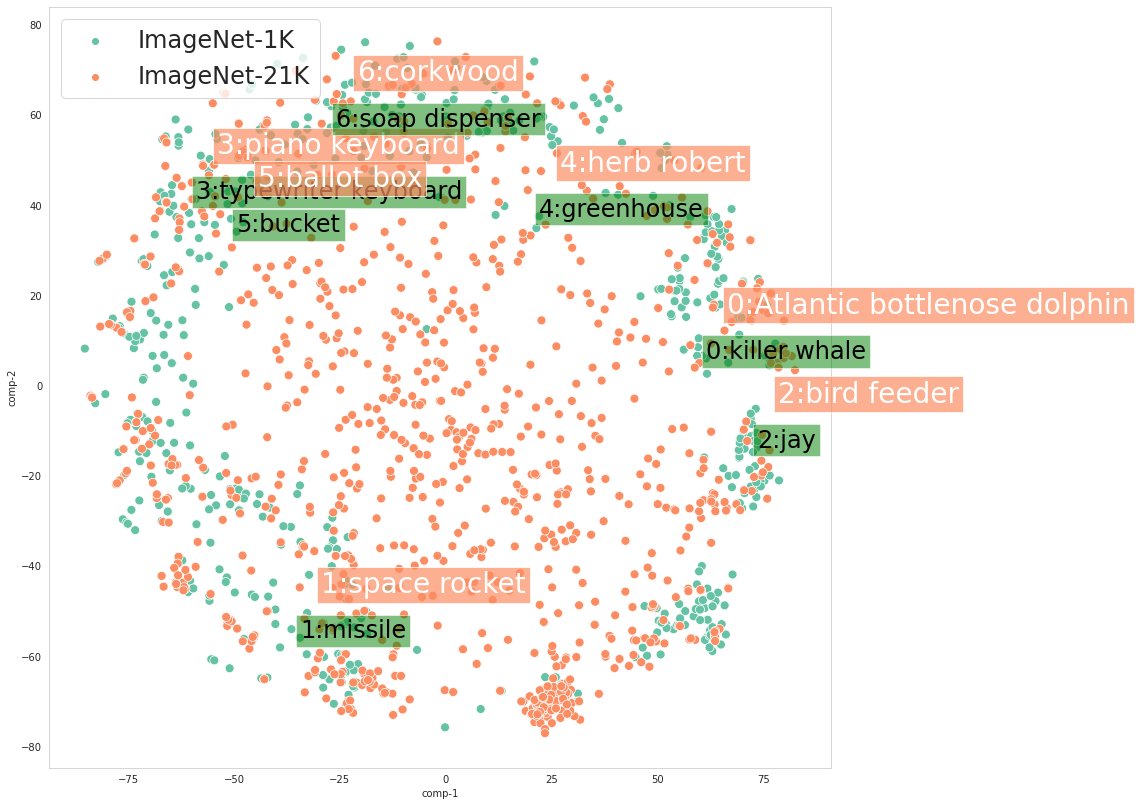}        
    \end{minipage}         
    \caption{Similar to Fig.~\ref{fig:concept_tsne}, we visualize the t-SNE embedding for another random set of visual concepts with models trained with ImageNet-1K (left) and ImageNet-1K+GCC-15M (right). Clearly, our model learned from the combination of image-label and image-text pairs can understand more number of visual concepts.}
    \label{fig:concept_tsne2}
    \vspace{-3mm}
\end{figure*}

In Fig.~\ref{fig:concept_tsne}, we further show the concept embeddings for two models as in Fig.~4 in our main submission. Fig.~\ref{fig:concept_tsne} left shows the model trained only on ImageNet-1K while right shows the model trained jointly with ImageNet-1K and GCC-15M. The model trained with two type of data understand the novel concepts from ImageNet-21K much better than the left one. For example, the left model put ``porthole'' and porcupine close to each other but the former is a circular window and latter is an animal. In contrast, the model at right side can easily find the ``porcuponefish'' as the close neighbor. Similarly, the left model mix ``goblet'' and ``coverlet'', probably because they share the same suffix. Our model on right side finds one of the most matched concepts ``liqueur glass'' which is semantically and visually similar to the query concept. Similar trend is also observed in Fig.~\ref{fig:concept_tsne2}. All these visualizations demonstrate that our model trained with both type of data has learned the visually-grounded semantic meanings for various concepts. 

\end{document}